\documentclass[10pt,twocolumn,letterpaper]{article}

\usepackage{iccv}
\usepackage{times}
\usepackage{epsfig}
\usepackage{graphicx}
\usepackage{amsmath}
\usepackage{amssymb}
\usepackage{booktabs}
\usepackage{url}            
\usepackage{amsfonts}       
\usepackage{nicefrac}       
\usepackage{microtype}      
\usepackage{xcolor}         
\usepackage{caption}
\usepackage{amsthm,amsmath,amssymb}
\usepackage{mathrsfs}
\usepackage{amssymb}
\usepackage{color}
\usepackage{multirow}
\usepackage{rotating}
\usepackage{algorithm}
\usepackage{listings}
\usepackage{longtable}
\usepackage{wrapfig,lipsum,booktabs}
\usepackage[noend]{algpseudocode}
\usepackage{colortbl}

\usepackage[pagebackref=true,breaklinks=true,letterpaper=true,colorlinks,bookmarks=false]{hyperref}

\iccvfinalcopy 

\def\iccvPaperID{5930} 

\ificcvfinal\fi

\begin{document}


\title{Improving Transformer-based Image Matching by Cascaded Capturing Spatially Informative Keypoints}


\author{Chenjie Cao, Yanwei Fu\footnotemark[2]\\
School of Data Science, Fudan University\\
{\tt\small \{20110980001,yanweifu\}@fudan.edu.cn}
}

\maketitle
\ificcvfinal\thispagestyle{empty}\fi

\renewcommand{\thefootnote}{\fnsymbol{footnote}}
\footnotetext[2]{ Corresponding author.}

\begin{abstract}
Learning robust local image feature matching is a fundamental low-level vision task, which has been widely explored in the past few years. 
Recently, detector-free local feature matchers based on transformers have shown promising results, which largely outperform pure Convolutional Neural Network (CNN) based ones.
But correlations produced by transformer-based methods are spatially limited to the center of source views' coarse patches, because of the costly attention learning.
In this work, we rethink this issue and find that such matching formulation degrades pose estimation, especially for low-resolution images.
So we propose a 
transformer-based cascade matching model -- Cascade feature Matching TRansformer (CasMTR)\footnote[4]{Code is available at \url{https://github.com/ewrfcas/CasMTR}}, to efficiently learn dense feature correlations, which allows us to choose more reliable matching pairs for the relative pose estimation.  
Instead of re-training a new detector, we use a simple yet effective Non-Maximum Suppression (NMS) post-process to filter keypoints through the confidence map, and largely improve the matching precision. CasMTR achieves state-of-the-art performance in indoor and outdoor pose estimation as well as visual localization. Moreover, thorough ablations show the efficacy of the proposed components and techniques.
\end{abstract}

\section{Introduction}
\label{sec:intro}

\begin{figure*}
\begin{centering}
\includegraphics[width=0.9\linewidth]{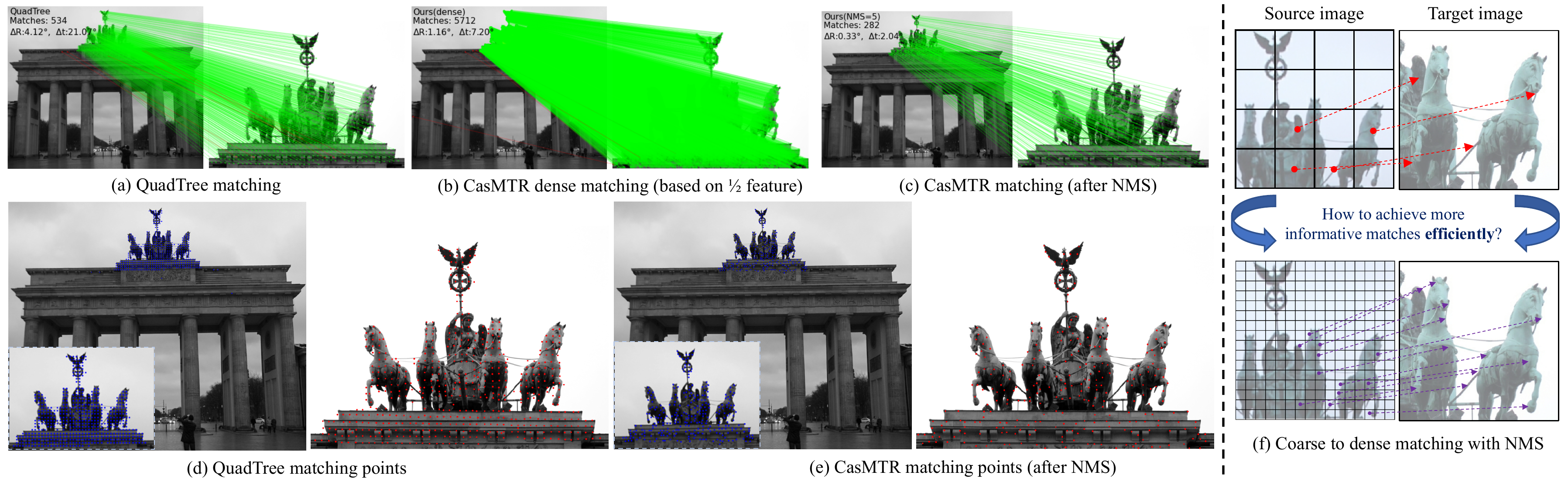} 
\par\end{centering}
\vspace{-0.1in}
\caption{QuadTree~\cite{tang2022quadtree} (a,d) vs our CasMTR (b,c,e). Our method achieves more fine-grained matching pairs for both source and target images (b). It is further improved by our NMS detection, which retains reliable matching results located in structural keypoints (c,e). We show an intuitive motivation for our spatially informative keypoints in (f). Best viewed in color.\label{fig:teaser}}
\vspace{-0.15in}
\end{figure*}

\begin{figure}
\begin{centering}
\includegraphics[width=0.75\linewidth]{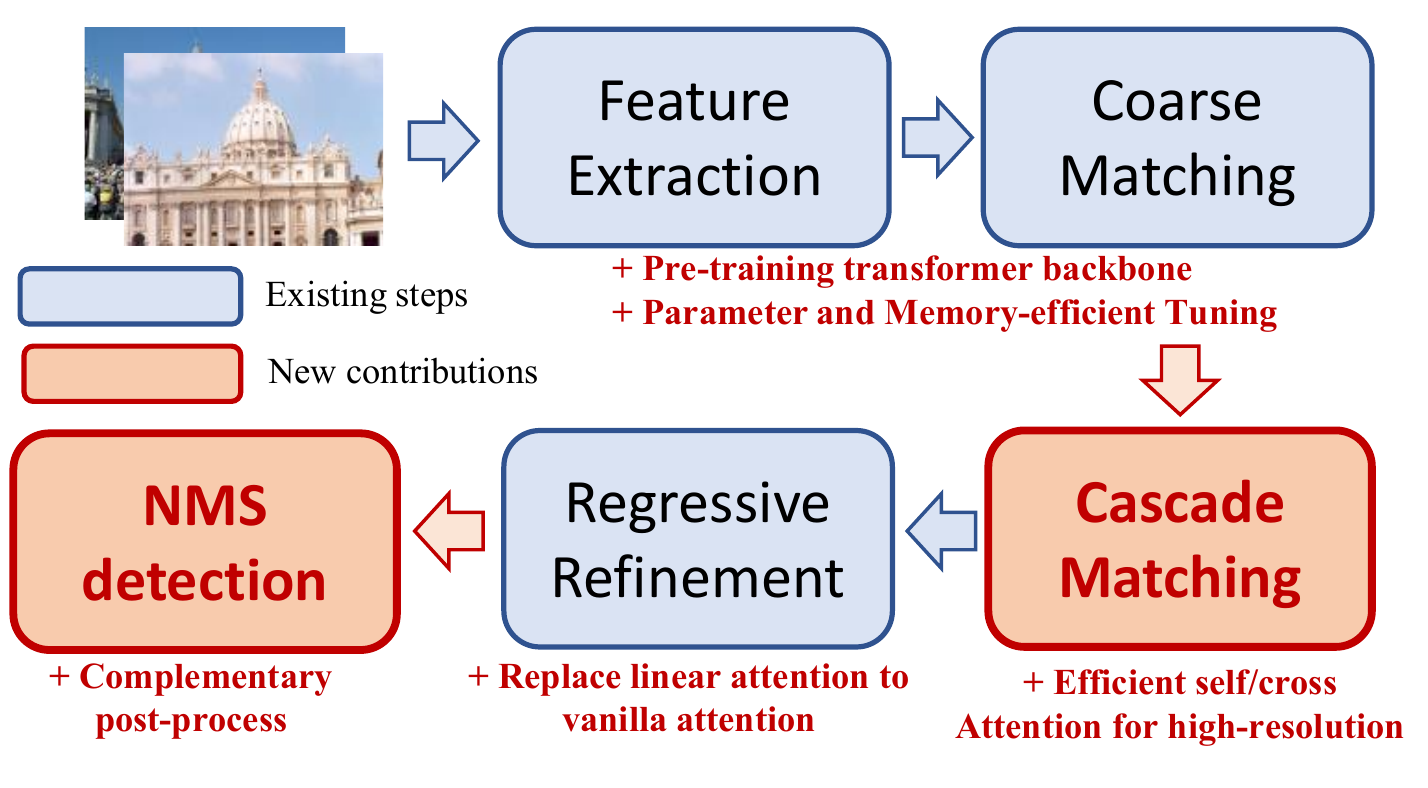} 
\par\end{centering}
\vspace{-0.15in}
\caption{Illustration of CasMTR pipeline; and our novelties compared against the existing steps from detector-free matching methods~\cite{sun2021loftr,tang2022quadtree,chen2022aspanformer} are highlighted in red.\label{fig:teaser_pipeline}
}
\vspace{-0.2in}
\end{figure}

\begin{table}
\footnotesize 
\caption{\small Summary of test image size, backbones, memory cost (GB), and inference speed (s/image) on MegaDepth~\cite{li2018megadepth} with AUC of different pose errors (\%). Suffixes `-8c', `-4c', and `-2c' denote matching at 1/8, 1/4, and 1/2 of image size. Baseline: QuadTree~\cite{tang2022quadtree} with the same backbone as ours.
Directly implementing QuadTree-4c causes Out-of-memory (OOM) error in a 32GB GPU, so its inference speed is estimated in brackets.
\label{tab:ablation_scale_backbone}}
\vspace{-0.1in}
\centering
\setlength{\tabcolsep}{0.7mm}{
\begin{tabular}{c|c|c|ccc|c|c}
\hline 
\multirow{2}{*}{Methods} & \multirow{2}{*}{Backbone} & \multirow{2}{*}{Size} & \multicolumn{3}{c|}{Pose Est. AUC} & \multirow{2}{*}{Mem.(G)} & \multirow{2}{*}{s/img}\tabularnewline
\cline{4-6} \cline{5-6} \cline{6-6} 
 &  &  & @5° & @10° & @20° &  & \tabularnewline
\hline 
Baseline-8c & Twins+FPN & 704 & 51.63 & 68.54 & 80.98 & 3.83 & 0.146\tabularnewline
CasMTR-4c & Twins+FPN & 704 & 52.59 & 69.78 & 82.31 & 3.99 & 0.212\tabularnewline
CasMTR-2c & Twins+FPN & 704 & \textbf{54.91} & \textbf{71.27} & \textbf{83.01} & 6.29 & 0.311\tabularnewline
\hline 
QuadTree-8c & FPN & 832 & 52.87 & 69.24 & 81.32 & 5.72 & 0.203\tabularnewline
Baseline-8c & Twins+FPN & 832 & 52.90 & 69.78 & 82.05 & 4.91 & 0.207\tabularnewline
QuadTree-4c & FPN & 832 & -- & -- & -- & OOM & (0.602)\tabularnewline
CasMTR-4c & Twins+FPN & 832 & 53.63 & 70.34 & 82.55 & 4.91 & 0.304\tabularnewline
CasMTR-2c & Twins+FPN & 832 & \textbf{55.61} & \textbf{71.96} & \textbf{83.52} & 7.55 & 0.444\tabularnewline
\hline 
QuadTree-8c & FPN & 1152 & 55.09 & 71.31 & 83.20 & 12.62 & 0.424\tabularnewline
Baseline-8c & Twins+FPN & 1152 & 55.77 & 72.01 & 83.64 & 13.33 & 0.423\tabularnewline
QuadTree-4c & FPN & 1152 & -- & -- & -- & OOM & (1.442)\tabularnewline
CasMTR-4c & Twins+FPN & 1152 & 56.34 & 72.11 & 83.55 & 12.40 & 0.649\tabularnewline
CasMTR-2c & Twins+FPN & 1152 & \textbf{56.90} & \textbf{72.94} & \textbf{84.24} & 14.36 & 0.887\tabularnewline
\hline 
\end{tabular}}
\vspace{-0.15in}
\end{table}

Image matching is an important vision problem that is widely employed for many downstream tasks like Structure-from-Motion~\cite{schonberger2016structure}, Simultaneous Localization and Mapping~\cite{mur2015orb}, and visual localization~\cite{lynen2020large}.
However, accurately matching two or more images remains difficult due to various factors, such as differences in viewpoints, illuminations, seasons, and surroundings.
Classical approaches~\cite{lowe2004distinctive,rublee2011orb} address it via the pipeline of \textit{detection, description, and matching of features} by hand-crafted features. 
Recently, learning Convolutional Neural Network (CNN) based detectors~\cite{detone2018superpoint,revaud2019r2d2,sarlin2020superglue,tian2020d2d,tyszkiewicz2020disk} have been utilized to detect and describe keypoints, leading to significant improvements in this pipeline.
But such detector-based CNNs suffer from limited receptive fields and search space, as noticed in~\cite{sun2021loftr}. 

To solve this issue, transformer-based detector-free methods have emerged as more robust alternatives, demonstrating impressive matching abilities in texture-less regions~\cite{sun2021loftr,jiang2021cotr,tang2022quadtree,wang2022matchformer,chen2022aspanformer}. 
However, the high computational cost of attention limits transformer-based methods to `\textit{semi-dense}' matching, where source matching points are spaced apart at intervals of coarse feature space, as shown in Fig.\ref{fig:teaser}(a,d).
Such semi-dense matching leads to an issue that \textit{keypoint locations are not informative enough}: the spatially restricted source points in coarse feature maps lack the necessary details to express structural information, making it difficult to accurately estimate pose. This problem is especially challenging for low-resolution images, as seen in our pilot study (Tab.~\ref{tab:ablation_scale_backbone}). More experiments based on extreme resolutions are discussed in the supplementary.
Furthermore, it remains unclear whether transformer-based methods can capture matching points in finer-grained image features rather than coarse ones (1/8) without a substantial increase in computational costs.

To address these challenges, we improve the existing transformer-based matching pipeline~\cite{sun2021loftr,chen2022aspanformer} by efficiently capturing spatially informative keypoints in a cascaded manner.
Particularly, our key idea is inspired by the coarse-to-fine Multi-View Stereo (MVS)~\cite{gu2020cascade}. We propose enhancing the transformer-based matching pipeline by adding the new stages of cascade matching and Non-Maximum Suppression (NMS) detection as summarized in Fig.~\ref{fig:teaser_pipeline}. Such new stages increase and refine the matching candidates in source views. Thus, we can achieve \emph{dense matching for both source and target views} as in Fig.~\ref{fig:teaser}(f), resulting in more precise matches focusing on more reliable positions with informative structures. Moreover, we elaborate on several novel techniques to support the newly incorporated stages in Fig.~\ref{fig:teaser_pipeline}. Consequently, the proposed method can achieve dense and precise matches on 1/2 image size located in informative space.

Formally, we propose a transformer-based matching method called Cascade feature Matching TRansformer (CasMTR).  It makes a significant contribution by enabling pure transformer-based models to conduct \textit{dense} matching by cascaded capturing spatially informative keypoints without relying on the expensive learning of huge 4D correlations as merely extended from~\cite{sun2021loftr}.
CasMTR develops several key components as follows.
Firstly, inherited in MVS, coarse-to-fine cascade matching modules are repurposed with different efficient attention mechanisms~\cite{katharopoulos2020transformers,wang2021pyramid,chu2021twins,tang2022quadtree,guo2022visual,zhao2022global} to overcome the semi-dense matching in coarse features. 
We present the local non-overlapping~\cite{chu2021twins} and overlapping~\cite{zhao2022global} self-attention for outdoor and indoor cases respectively, due to different illuminations and surroundings.
Secondly, CasMTR enjoys flexible training by a novel Parameter and Memory-efficient Tuning method (PMT), which is originally derived for NLP tasks~\cite{sung2022lst}. 
Essentially, PMT can incrementally finetune CasMTR based on off-the-shelf matching models with reliable coarse matching initialization and fast convergence.
Thirdly, we for the first time introduce the training-free NMS detection to complementarily filter precise matches based on dense matching confidence maps of CasMTR.
Critically, NMS serves as a simple yet effective post-processing that preserves structurally meaningful keypoints rather than the coarse patch center as in Fig.~\ref{fig:teaser}(e). This improves the pose estimation as in Fig.~\ref{fig:teaser}(c) and has good generalization for various high-resolution matching tasks~\cite{li2018megadepth,balntas2017hpatches,taira2018inloc,zhang2021reference}.
Finally, in the development of our model, we have learned that the devil is in the details. Consequently, several non-trivial technical improvements have been implemented to our newly proposed matching pipeline (highlighted in Fig.~\ref{fig:teaser_pipeline}), such as pre-training transformer backbones, improving efficient linear attention, and optimizing self and cross attention for high-resolution matching.


The proposed CasMTR is comprehensively evaluated in relative pose estimation~\cite{li2018megadepth,dai2017scannet}, homography estimation~\cite{balntas2017hpatches}, and visual localization~\cite{taira2018inloc,zhang2021reference}, showing its state-of-the-art performance. Additionally, our exhaustive ablation studies show the effectiveness of all newly proposed components. 



\section{Related Work}

\noindent\textbf{Detector-based Matching.}
Detector-based matching methods following the process of feature detection, description, and matching have dominated this field for a long time. Traditional manners utilize heuristic hand-craft features~\cite{lowe2004distinctive,rublee2011orb} for local feature matching, which enjoy great success and are still used in many 3D tasks nowadays.
After the wave of deep learning, many learning-based methods~\cite{han2015matchnet,yi2016lift,detone2017toward,dusmanu2019d2,liu2019gift,luo2020aslfeat} were proposed based on the detector-dependent pipeline with better performance. SuperPoint~\cite{detone2018superpoint} proposes to utilize the homographic adaptation for the self-supervised matching training. Then, SuperGlue~\cite{sarlin2020superglue} further improves the performance through the graph neural network. Moreover, DISK~\cite{tyszkiewicz2020disk} leverages reinforcement learning to optimize the end-to-end detector-based pipeline. However, these methods still suffer from limited interest points in indistinctive regions~\cite{sun2021loftr}. 
On the other hand, D2D~\cite{tian2020d2d} proposes to describe first, and then detect based on deep descriptors~\cite{mishchuk2017working,tian2019sosnet}. 
Compared with confidence-based NMS detection, feature-based D2D is complicated and slower. Besides, D2D is not compatible with the joint training model because it ignores the correlation between source and target views (Tab.~\ref{tab:ablation_detection}).

\noindent\textbf{Detector-free Matching.}
Detector-free methods enjoy an end-to-end pipeline to achieve the matching directly without an explicit keypoint detection phase~\cite{liu2010sift,choy2016universal,sun2021loftr}. Learning-based detector-free methods can be generally categorized into Convolutional Neural Network (CNN) based methods~\cite{rocco2018neighbourhood,rocco2020efficient,li2020dual,truong2020glu,truong2021learning,edstedt2022deep} and transformer or attention-based ones~\cite{sun2021loftr,jiang2021cotr,tang2022quadtree,wang2022matchformer,shi2022clustergnn,chen2022aspanformer,tan2022eco}. CNN-based methods produce dense matching results through learning 4D cost volumes or warped features, which are limited by receptive fields. 
Some transformer-based manners~\cite{sun2021loftr,tang2022quadtree,wang2022matchformer,chen2022aspanformer}, {led by LoFTR~\cite{sun2021loftr},} largely enlarge the receptive fields with interlacing self/cross attention modules, {and enjoy better performance in texture-less regions.
On the other hand, COTR~\cite{jiang2021cotr} jointly learns both matching images with self-attention together rather than modeling self/cross ones respectively in encoders. Then query points are decoded through cross-attention for the matching results. We focus on the former one in this paper.}
But matching {density and accuracy} of these approaches are insufficient to tackle many downstream tasks precisely, \textit{e.g.}, pose estimation for low-resolution images. Moreover, it is non-trivial to extend these transformer-based matching solutions into dense and high-resolution cases because of the heavy computation of attention.

\noindent\textbf{Coarse-to-fine Learning.}
The efficient coarse-to-fine manner plays an important role in learning-based stereo matching~\cite{tonioni2019real,wang2019anytime,yin2019hierarchical,gu2020cascade}, MVS~\cite{gu2020cascade,zhang2020visibility,wang2021patchmatchnet,mi2022generalized}, and optical flow~\cite{ranjan2017optical,sun2018pwc,yang2019volumetric,zhao2020maskflownet}. CasMVSNet~\cite{gu2020cascade} builds coarse cost volume at early stages with large depth ranges and makes later stages refine details. On the other hand, PWC-Net~\cite{sun2018pwc} warps pyramid features into cost volumes to further refine the coarse-to-fine flow estimation. 
Different from the depth prediction and the optical flow, learning geometric image matching with coarse-to-fine manners is more solid to tackle the error propagation~\cite{teed2020raft}. Because the geometric image matching is usually based on static landmarks with consistent displacements. Thus the coarse matching in low-resolution will not inevitably mislead local details.
Patch2pix~\cite{zhou2021patch2pix} proposed a coarse-to-fine refinement for pixel-level matching {just for CNNs.
COTR~\cite{jiang2021cotr} needs to recursively crop finer patches for more precise matching results, which is very time-consuming. ECO-TR~\cite{tan2022eco} proposes to crop coarse-to-fine feature patches and train them end-to-end to improve efficiency. But the feature cropping of ECO-TR still limits the receptive fields across different patches. 
Hence learning a coarse-to-fine transformer-based matching model with global receptive fields is still challenging.
}

\section{Method}

\begin{figure*}
\begin{centering}
\includegraphics[width=0.8\linewidth]{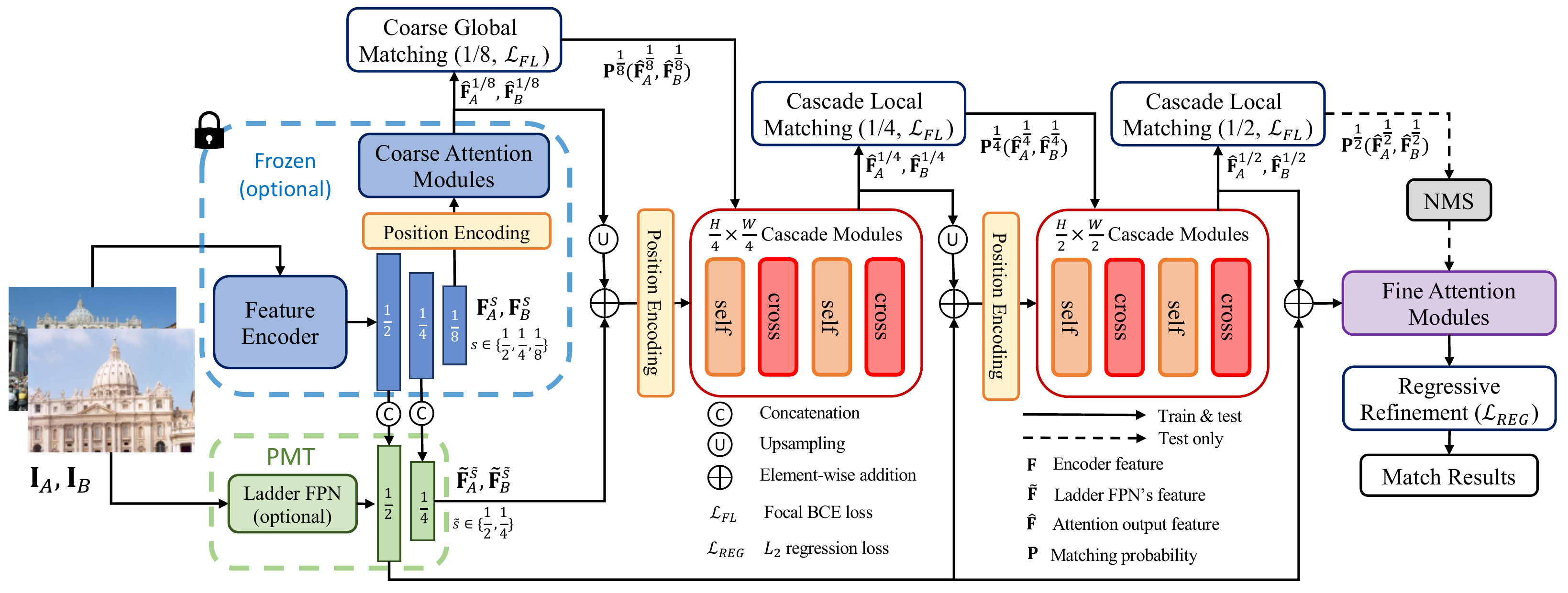} 
\par\end{centering}
\vspace{-0.1in}
\caption{Overview of CasMTR. Optionally, our model can work as an incremental refinement. Particularly, we could freeze feature encoder and coarse attention modules during training with a lightweight trainable ladder FPN to save the computation and memory footprint.
Matching scales and loss functions are denoted in the bracket of each matching module, \textcolor{black}{while feature scales are shown in superscripts.
Softmax matching probabilities $\mathbf{P}(\hat{\mathbf{F}}_A,\hat{\mathbf{F}}_B)$ got from global (1/8) and local (1/4, 1/2 detailed in Eq.~\ref{eq:similarity}) dot products are utilized to decide the next local matching candidates and NMS (test only).}
\label{fig:overview}}

\vspace{-0.15in}
\end{figure*}

\noindent\textbf{Preliminary and Overview.} 
We briefly review the transformer-based matching baseline in the example of LoFTR~\cite{sun2021loftr}. LoFTR uses a local feature CNN to extract coarse (1/8) and fine (1/2) feature maps from image pairs. Then interlaced self/cross-attention modules are leveraged to learn coarse-level matching predictions. Additionally, LoFTR utilizes a refinement module to model sub-pixel match prediction in fine-level features. However, the source point of each matched pair is still restricted at the coarse level (1/8), which limits the performance. Some follow-ups~\cite{tang2022quadtree,chen2022aspanformer} improve the linear attention~\cite{katharopoulos2020transformers} of LoFTR while retaining the whole pipeline unchanged.
Inherited from the LoFTR, we develop a novel coarse-to-fine CasMTR as in Fig.~\ref{fig:overview}.
Given the matching image pair $\mathbf{I}_A, \mathbf{I}_B$, we first extract their multi-scale features by a feature encoder. Then the self and cross QuadTree attention~\cite{tang2022quadtree} based coarse matching is performed in coarse-level features (Sec.~\ref{sec:feature_extraction}). 
According to the coarse matches, a couple of local attention-based cascade matching modules are proposed to refine the matching pairs (Sec.~\ref{sec:cascade_matching}). After that, a sub-pixel refinement leverages the spatial expectation to predict exact matching results (Sec.~\ref{sec:refinement}). Finally, the NMS post-process detects local keypoints based on confidence maps, which largely improves the pose estimation in outdoor scenes (Sec.~\ref{sec:nms_posprocess}). 

\subsection{Feature Extraction and Coarse Matching}
\label{sec:feature_extraction}

\noindent\textbf{Feature Encoder.} 
We first follow~\cite{sun2021loftr} and use FPN to produce coarse-to-fine features $\mathbf{F}_A^s,\mathbf{F}_B^s$ for the image pair $\mathbf{I}_A,\mathbf{I}_B$, where $s\in\{\frac{1}{2},\frac{1}{4},\frac{1}{8}\}$ indicate the image scale. Inspired by~\cite{huang2022flowformer}, we try to replace $\{\frac{1}{4},\frac{1}{8}\}$ layers with partial pre-trained Twins~\cite{chu2021twins} layers. To balance the computation, the feature encoder's channels are reduced in our CasMTR, which also benefits the efficiency of subsequent cascade modules. The pre-trained attention-based encoder strengthens the matching learning as evaluated in Tab.~\ref{tab:ablation_scale_backbone}.

\noindent\textbf{Parameter and Memory-efficient Tuning (PMT).} 
Since we pay more attention to the coarse-to-fine matching, our coarse matching is simply based on the state-of-the-art QuadTree attention~\cite{tang2022quadtree}. 
Essentially, the proposed model is learned independently from the coarse matching, \textit{i.e.}, we can freeze the feature encoder and coarse matching module, and incrementally finetune the cascade matching with the coarse matching initialization. To improve the representation of high-level features, we introduce PMT to incrementally finetune the matching model as shown in Fig.~\ref{fig:overview}. Specifically, a lightweight trainable ladder side FPN is utilized to receive and concatenate features from the frozen feature encoder as $\tilde{\mathbf{F}}_A^{\tilde{s}},\tilde{\mathbf{F}}_B^{\tilde{s}}$, where $\tilde{s}\in\{\frac{1}{2},\frac{1}{4}\}$. 
Different from other tuning techniques~\cite{houlsby2019parameter,li2021prefix,dong2022incremental}, PMT is not only parameter-efficient but also memory-efficient. 
Because the FPN of PMT could be well-updated by fine-grained features without any gradients propagated back from frozen models. 
In practice, we leverage the PMT to finetune our cascade modules based on the off-the-shelf QuadTree matching on the large ScanNet dataset~\cite{dai2017scannet}. Our algorithm can be converged in about two epochs and achieve appreciable improvements as in Tab.~\ref{tab:scannet}.


\subsection{Cascade Matching Modules}
\label{sec:cascade_matching}

Following the coarse matching results, we additionally propose multi-stage cascade modules to further refine more detailed matching results for both source and target images. For each stage, we first add sinusoidal position encoding as other methods~\cite{sun2021loftr,tang2022quadtree,chen2022aspanformer}. We normalize the position encoding as~\cite{chen2022aspanformer} during the inference, which makes CasMTR robust to various test sizes.
Then, self and cross-attention layers are interleaved in the cascade module for better local feature learning. 
Different from 1D-cascade architectures~\cite{gu2020cascade}, extending the cascade mechanism to 2D is non-trivial.
The main concern about cascade matching learning is the computation for high-resolution features. To address this, we thoroughly compare various efficient self and cross-attention mechanisms and choose the best combination among them. 

\begin{figure}
\begin{centering}
\includegraphics[width=0.9\linewidth]{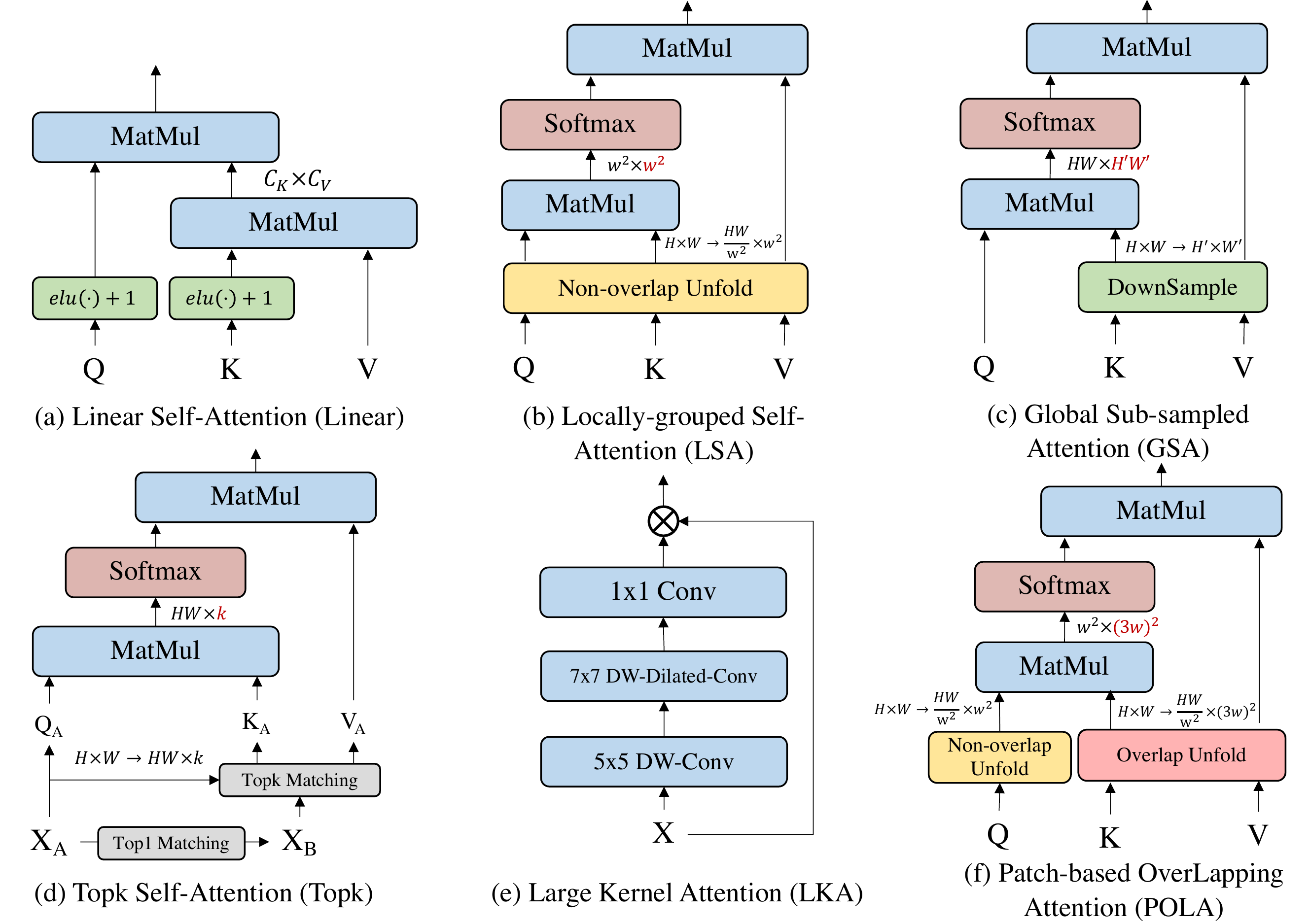} 
\par\end{centering}
\vspace{-0.1in}
\caption{Six self-attention modules explored in CasMTR.\label{fig:self_attns}}
\vspace{-0.15in}
\end{figure}

\noindent\textbf{Self-Attention.}  
Global self-attention suffers from quadratic spatial complexity, especially for high-resolution features. 
But without the self-attention, the pure cross-attention model performs not well as in Tab.~\ref{tab:ablation_attns}.
To balance the computation and the performance, we explore six efficient self-attention mechanisms shown in Fig.~\ref{fig:self_attns} and verified in Tab.~\ref{tab:ablation_attns}, including Linear attention~\cite{katharopoulos2020transformers}, Locally-grouped Self-Attention (LSA)~\cite{chu2021twins}, Global Sub-sampled Attention (GSA)~\cite{wang2021pyramid}, simplified top-k attention, Large Kernel Attention (LKA)~\cite{guo2022visual}, and Patch-based OverLapping Attention (POLA)~\cite{zhao2022global}. 
We list more details of these manners in the supplementary.

\begin{table}
\footnotesize 
\caption{Pilot study of AUC and FLOPs about different attention mechanisms based on 1/4 cascade model (Ours-4c) on MegaDepth.  All FLOPs of cascade modules are based on 1152$\times$1152 test images. \textcolor{red!75}{LSA+LW} is adopted on MegaDepth, while \textcolor{blue!75}{POLA+LW} is adopted on ScanNet. \label{tab:ablation_attns}}
\vspace{-0.1in}
\centering
\setlength{\tabcolsep}{0.5mm}{
\begin{tabular}{c|c|ccc|ccc|c}
\hline 
\multicolumn{2}{c|}{Cascade (layers)} & \multicolumn{3}{c|}{MegaDepth} & \multicolumn{3}{c|}{ScanNet} & \multirow{2}{*}{FLOPs(G)}\tabularnewline
\cline{1-8} \cline{2-8} \cline{3-8} \cline{4-8} \cline{5-8} \cline{6-8} \cline{7-8} \cline{8-8} 
self & cross & @5$^{\circ}$ & @10$^{\circ}$ & @20$^{\circ}$ & @5$^{\circ}$ & @10$^{\circ}$ & @20$^{\circ}$ & \tabularnewline
\hline 
Linear(2) & LW(2) & 56.01 & 72.03 & 83.43 & -- & -- & -- & 129.21\tabularnewline
\cellcolor{red!25}LSA(2) & \cellcolor{red!25}LW(2) & \cellcolor{red!25}56.34 & \cellcolor{red!25}72.11 & \cellcolor{red!25}83.55 & \cellcolor{red!15}26.24 & \cellcolor{red!15}46.45 & \cellcolor{red!15}63.94 & \cellcolor{red!25}142.32 \tabularnewline
\scriptsize{}{LSA+GSA(2)} & LW(2) & 55.71 & 71.60 & 83.17 & -- & -- & -- & 202.56 \tabularnewline
LSA(2) & MT(2) & 55.60 & 71.92 & 83.27 & -- & -- & -- & 143.51 \tabularnewline
LKA(2) & LW(2) & 55.75 & 72.02 & 83.20 & -- & -- & -- & 136.53 \tabularnewline
-- & LW(4) & 55.16 & 71.48 & 83.01 & -- & -- & -- & 143.28 \tabularnewline
Top-k(2) & LW(2) & 55.47 & 71.28 & 83.02 & -- & -- & -- & 141.76 \tabularnewline
LKA(2) & MT(2) & \textbf{56.99} & \textbf{72.56} & \textbf{83.89} & 25.79 & 45.87 & 63.50 & 137.72 \tabularnewline
\cellcolor{blue!25}POLA(2) & \cellcolor{blue!25}LW(2) & \cellcolor{blue!15}56.31 & \cellcolor{blue!15}72.35 & \cellcolor{blue!15}83.51 & \cellcolor{blue!25}\textbf{27.08} & \cellcolor{blue!25}\textbf{47.02} & \cellcolor{blue!25}\textbf{64.44} & \cellcolor{blue!25}223.42\tabularnewline
\hline 
\end{tabular}}
\vspace{-0.15in}
\end{table}

\begin{figure}
\begin{centering}
\includegraphics[width=0.9\linewidth]{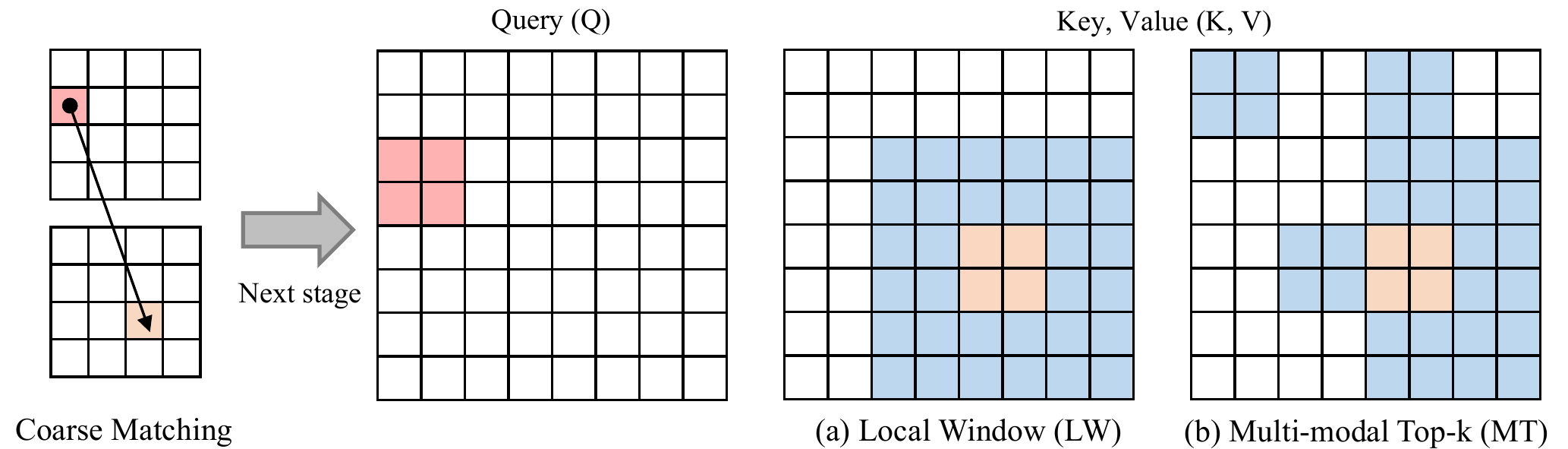} 
\par\end{centering}
\vspace{-0.1in}
\caption{Two cross-attentions explored in cascade modules.\label{fig:cross_attns}}
\vspace{-0.2in}
\end{figure}

\noindent\textbf{Cross-Attention.}
The cross-attention plays an important role in cascade matching. Two types of attention modules, designed in Fig.~\ref{fig:cross_attns} are Local Window (LW) and Multi-modal Top-k (MT) cross-attention respectively.
Given the coarse matching result, each query patch in LW intuitively selects neighbor patches around the top-1 matching target from another image as key and value patches. In Fig.~\ref{fig:cross_attns}(a), the LW example is based on window size 6 and 36 neighbors in all. LW dramatically reduces the sequence length of keys and values, which makes the cascade matching for high-resolution features possible. Furthermore, LW is more capable of learning detailed feature correlations.
On the other hand, to alleviate the intractable error propagation caused by the coarse stage~\cite{yang2022non}, we propose MT to model multi-modal distribution for cascade matching. In particular, MT holds top-k coarse matching patches as candidates. Then they are upsampled as key and value patches for the cross-attention. 
As in Fig.~\ref{fig:cross_attns}(b), the MT example is on top-36: top-9 in the coarse stage, and each coarse block can be further divided into 4 patches in the cascade stage. MT can potentially address the mismatch from the coarse stage, as long as the top-k candidates can cover the ground truth. 
Because of the scale upsampling, both source and target features are quadrupled as shown in Fig.~\ref{fig:cross_attns}, which influence the practical kernel size and top-k in LW and MT respectively. 
Technically, we implement both LW and MT by CUDA to improve efficiency, and their speed is almost the same in practice.

\noindent\textbf{Analysis of Attention Modules.}
We conduct pilot study to verify the results and FLOPs of all attention combinations in Tab.~\ref{tab:ablation_attns}. Such a pilot thus guides how we  design the  model. 
Specifically, `LSA+LW' enjoys a good trade-off between performance and efficiency in the outdoor MegaDepth. Furthermore, the extended `LSA+GSA' fails to achieve better results with more computation. We think that local feature learning is more important than global one in our cascade modules.
Besides, for the indoor scenes from ScanNet with more challenging texture-less matching instances, `POLA+LW' outperforms `LSA+LW' with larger receptive fields.
Therefore, `LSA+LW' and `POLA+LW' are used to comprise our cascade modules.
Note that `LKA+MT' can produce superior results in MegaDepth. But we did not choose `LKA+MT' for two reasons. First, the depthwise convolutions used in LKA are not well optimized in PyTorch, which largely slows down the training. Second, `LKA+MT' is not stable enough, as it fails to achieve reliable results in ScanNet.

\noindent\textbf{Matching and Loss.}
Given $\hat{\mathbf{F}}^{\tilde{s}}_A,\hat{\mathbf{F}}^{\tilde{s}}_B, \tilde{s}\in\{\frac{1}{2},\frac{1}{4}\}$ after the interlaced attention learning of cascade modules, we use the same key candidates from the cross-attention (\textit{i.e.}, LW or MT) for the dot product similarity matrix as
\begin{equation}
\label{eq:similarity}
\mathbf{S}(i,j)=\frac{1}{\tau}\cdot\left\langle\hat{\mathbf{F}}^{\tilde{s}}_A(i),\hat{\mathbf{F}}^{\tilde{s}}_B(j)\right\rangle\in\mathbb{R}^{H^{\tilde{s}}W^{\tilde{s}}\times k}
\end{equation}
where $\tau=0.1$ is a scale parameter; the key length $k$ is 100 and 128 for LW and MT respectively. 
Note that Eq.~\ref{eq:similarity} presents a local correlation with $k$ candidates for each feature point rather than the full correlation in the coarse level.
Softmax is used to normalize Eq.~\ref{eq:similarity} into local matching probability $\mathbf{P}^{\tilde{s}}(i,j)$. 
We also adopt cycle-consistent matching to enforce that two features in different images are matched each other.
Following~\cite{sun2021loftr}, the Focal binary cross-entropy Loss (FL)~\cite{lin2017focal} is used to optimize the cascade matching as
\begin{equation}
\label{eq:cascade_loss}
\mathcal{L}^{\tilde{s}}_{FL}=-\mathbb{E}_{\mathcal{M}^{\tilde{s}}}[(1-\mathbf{P}^{\tilde{s}})^{\gamma}\log(\mathbf{P}^{\tilde{s}})],
\end{equation}
where $\gamma=2$; $\mathcal{M}^{\tilde{s}}$ indicates matching queries which satisfy the cycle-consistent and have one ground truth target in $k$ candidates. 
In cascade stages, the classification loss enjoys the priority because we have to learn proper confidence~\cite{cao2022mvsformer} for the detection (Sec.~\ref{sec:nms_posprocess}). 
We also tried the vanilla cross-entropy, but it performed slightly worse than FL.

\noindent\textbf{Discussions.} 
Since cascade matching facilitates pose estimation in limited resolution (Tab.~\ref{tab:ablation_scale_backbone}), our method achieves prominent improvement on 480$\times$640 ScanNet~\cite{dai2017scannet}  without any post-processing (Tab.~\ref{tab:scannet}). When input images become larger, the coarse matching pairs also become dense gradually to alleviate the pose estimation error. 
Moreover, for large image scales, we find that our cascade matching can be strengthened by a simple yet effective NMS post-processing with negligible cost.
Besides, our CasMTR is efficient enough compared with the trivial extension (QuadTree-4c).

\subsection{Local Regressive Refinement}
\label{sec:refinement}
The patch-wise refinement module in LoFTR~\cite{sun2021loftr} is also incorporated in our work for sub-pixel matching. The refinement module first unfolds all features into $5\times5$ patches. Different from the one in LoFTR, we use the standard attention instead of the linear one in both self and cross attention. Because the refinement module only calculates the attention map with a sequence length $5\times5=25$. Therefore, standard attention even enjoys less computation compared with linear attention and performs better. The refinement module utilizes soft-argmax to regress the residual matching flow. One may ask whether such local refinement can replace cascade modules for dense matching. We should clarify that the patch-wise refinement is extremely limited by the matching range and receptive fields, which discourages the results. We tried the trivial solution to densely match through the refinement module in Tab.~\ref{tab:trivial_solution}, but it worked worse than the baseline. Even NMS failed to make its results competitive.

\subsection{Confidence based Detection with NMS}
\label{sec:nms_posprocess}

Different from detector-based methods~\cite{dusmanu2019d2,detone2018superpoint,sarlin2020superglue}, latest attention based methods~\cite{sun2021loftr,tang2022quadtree,chen2022aspanformer,wang2022matchformer} achieve good results even  without detector. These detector-free methods only use a confidence threshold to filter unconvinced matching. Moreover, the sparse matching (1/8) is not ready for further keypoint detection. Except for the confidence threshold, we propose to use the simple NMS to detect local keypoints through the cascade confidence maps as shown in Fig.~\ref{fig:teaser}(c). 
Specifically, we apply the overlapping max-pooling on the confidence map. Then if the local maximal confidence locates in the center of the pooling kernel, we retain this matching pair. 
So the minimum interval in feature space of two keypoints is equivalent to half of NMS's kernel size.
The main difference between the NMS and the threshold refusing is that NMS detects local keypoints through confidence rather than global filtering. Thanks to the dense matching from cascade modules, NMS can shift the matching prediction to some structural keypoints with relatively higher confidence. Thus NMS is complementary to CasMTR.
We find that the simple NMS outperforms other traditional detectors~\cite{lowe2004distinctive}, and feature-based detector~\cite{dusmanu2019d2,tian2020d2d}. Further, we train CasMTR with trainable detectors, working worse than NMS as in Tab.~\ref{tab:ablation_detection}, which is discussed in Sec.~\ref{sec:abaltions}. 

\section{Experiments}


\noindent\textbf{Datasets.}
CasMTRs are trained on outdoor MegaDepth~\cite{li2018megadepth} and indoor ScanNet~\cite{dai2017scannet} to verify the relative pose estimation. MegaDepth comprises 196 scene reconstructions with 1M Internet images. Ground-truth matching pairs are from COLMAP~\cite{schonberger2016pixelwise} computed depth maps, Following~\cite{sun2021loftr}, for one epoch, we randomly sample 200 pairs from each scene for the training, and 1500 pairs from independent two scenes are selected as the test set. For the ScanNet, there are 1613 monocular sequences with 230M and 1500 pairs for training and testing respectively. For one epoch, 100 images are sampled for training on each scene.

\noindent\textbf{Implementation.}
We extend CasMTR into $\{\frac{1}{4},\frac{1}{2}\}$ resolutions with cascade modules, \textit{i.e.}, CasMTR-4c and CasMTR-2c. The NMS kernel is fixed in 5 for the pose estimation. For MegaDepth, CasMTR is trained progressively in $704\times704$ and tested in $1152\times1152$. In particular, we first train CasMTR in the coarse stage with $\frac{1}{8}$ matching for 8 epochs. Then CasMTR-4c and CasMTR-2c are further finetuned with 16 and 8 epochs respectively. \textcolor{black}{CasMTR-2c converges faster than CasMTR-4c because more supervised matching pairs are learned in the high-resolution learning for each epoch.} For ScanNet, both training and testing image size is $480\times640$. To tackle the mega data scale, we use PMT to incrementally finetune CasMTR-4c based on the off-the-shelf QuadTree~\cite{tang2022quadtree} weights. PMT-CasMTR-4c can converge in only 2 epochs. CasMTR shares a 0.2 threshold in all stages.


\subsection{Relative Pose Estimation}
As in~\cite{sarlin2020superglue,sun2021loftr}, we evaluate the relative pose estimation with AUC of pose errors at thresholds ($5^\circ, 10^\circ, 20^\circ$), while the pose error is defined as the maximum angular error of rotation and translation. The essential matrix is optimized by OpenCV RANSAC with model-predicted matching pairs.

\begin{figure*}
\begin{centering}
\includegraphics[width=0.9\linewidth]{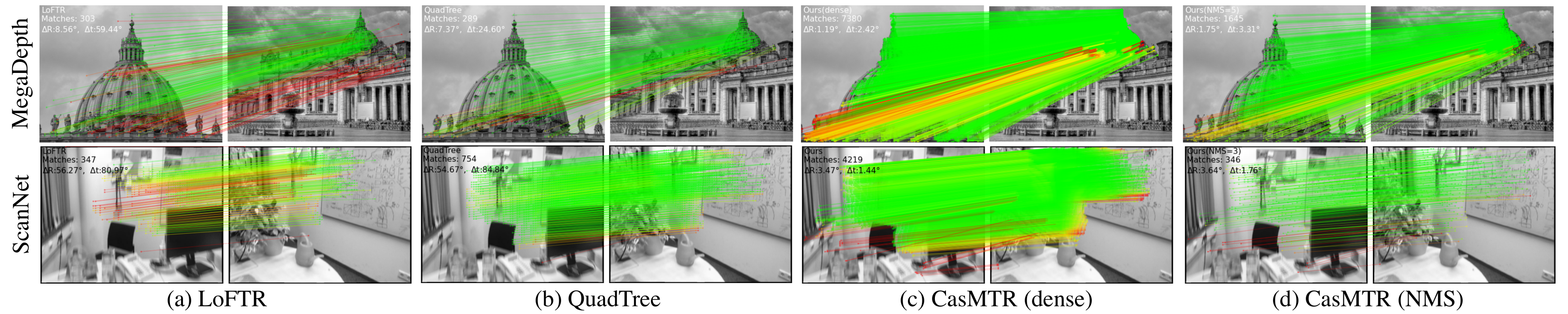} 
\par\end{centering}
\vspace{-0.1in}
\caption{Qualitative outdoor and indoor matching results compared with LoFTR~\cite{sun2021loftr}, QuadTree~\cite{tang2022quadtree}, CasMTR-4c (ScanNet), CasMTR-2c (MegaDepth), and our NMS detected results.\label{fig:qualitative}}
\vspace{-0.15in}
\end{figure*}

\noindent\textbf{Outdoor MegaDepth.} 
We show MegaDepth results in Tab.~\ref{tab:megadepth}. CasMTR can outperform all competitors especially in AUC$5^\circ$ and AUC$10^\circ$, which include both transformer-based~\cite{sun2021loftr,tang2022quadtree,wang2022matchformer,chen2022aspanformer} and CNN-based~\cite{edstedt2022deep} manners. Moreover, our CasMTR-2c with denser feature matching capability can further improve the performance.
Besides, our NMS detection is effective for outdoor scenes with large displacements and appearance transformations as verified in Tab.~\ref{tab:ablation_detection}.
Qualitative results are compared in Fig.~\ref{fig:qualitative}. Our CasMTR achieves denser and more exact matching results.

\begin{table}
\small 
\caption{Pose estimation on outdoor MegaDepth with AUC of different pose errors (\%). 
\label{tab:megadepth}}
\vspace{-0.1in}
\centering
\begin{tabular}{c|ccc}
\hline 
\multirow{2}{*}{Methods} & \multicolumn{3}{c}{Pose Estimation AUC $\uparrow$}\tabularnewline
\cline{2-4} \cline{3-4} \cline{4-4} 
 & @5$^\circ$ & @10$^\circ$ & @20$^\circ$\tabularnewline
\hline
SP~\cite{detone2018superpoint}+SuperGlue~\cite{sarlin2020superglue} & 42.2 & 61.2 & 76.0\tabularnewline
\hline
PDCNet+(H)~\cite{truong2021pdc} & 43.1 & 61.9 & 76.1\tabularnewline
LoFTR~\cite{sun2021loftr} & 52.8 & 69.2 & 81.2\tabularnewline
QuadTree~\cite{tang2022quadtree} & 54.6 & 70.5 & 82.2\tabularnewline
MatchFormer~\cite{wang2022matchformer} & 53.3 & 69.7 & 81.8\tabularnewline
DKM~\cite{edstedt2022deep} & 54.5 & 70.7 & 82.3\tabularnewline
ASpanFormer~\cite{chen2022aspanformer} & 55.3 & 71.5 & 83.1\tabularnewline
\hline
\textbf{CasMTR-4c} & 58.0 & 73.6 & 84.6\tabularnewline
\textbf{CasMTR-2c} & \textbf{59.1} & \textbf{74.3} & \textbf{84.8}\tabularnewline
\hline
\end{tabular}
\vspace{-0.15in}
\end{table}

\noindent\textbf{Indoor ScanNet.}
ScanNet results are in Tab.~\ref{tab:scannet}. CasMTR-4c achieves the best result among all competitors. Note that our PMT-enhanced method only needs to be finetuned with 2 epochs, which is flexible and efficient for the practice. CasMTR-2c did not obtain more improvement compared with CasMTR-4c in ScanNet. We think that texture-less regions of indoor scenes with motion blur and inferior annotations are too challenging for local attention learning in the 1/2 resolution.
Since the resolution of ScanNet is much lower than MegaDepth, NMS is not applied to CasMTR to remain dense enough matching pairs, which results in more precise pose estimation as qualitatively compared in Fig.~\ref{fig:qualitative}.

\begin{table}
\small 
\caption{Pose estimation on indoor ScanNet~\cite{dai2017scannet} with AUC of different pose errors (\%).
\label{tab:scannet}}
\vspace{-0.1in}
\centering
\begin{tabular}{c|ccc}
\hline
\multirow{2}{*}{Methods} & \multicolumn{3}{c}{Pose Estimation AUC $\uparrow$}\tabularnewline
\cline{2-4} \cline{3-4} \cline{4-4} 
 & @5$^\circ$ & @10$^\circ$ & @20$^\circ$\tabularnewline
\hline
SP~\cite{detone2018superpoint}+SuperGlue~\cite{sarlin2020superglue} & 16.2 & 33.8 & 51.8\tabularnewline
\hline
PDCNet+(H)~\cite{truong2021pdc} & 20.2 & 39.4 & 57.1\tabularnewline
LoFTR~\cite{sun2021loftr} & 22.0 & 40.8 & 57.6\tabularnewline
QuadTree~\cite{tang2022quadtree} & 24.9 & 44.7 & 61.8\tabularnewline
MatchFormer~\cite{wang2022matchformer} & 24.3 & 43.9 & 61.4\tabularnewline
DKM~\cite{edstedt2022deep} & 24.8 & 44.4 & 61.9\tabularnewline
ASpanFormer~\cite{chen2022aspanformer} & 25.6 & 46.0 & 63.3\tabularnewline
\hline
\textbf{CasMTR-4c} & \textbf{27.1} & \textbf{47.0} & \textbf{64.4}\tabularnewline
\hline
\end{tabular}
\vspace{-0.15in}
\end{table}

\subsection{Homography Estimation}
CasMTR is also evaluated in on HPatches dataset~\cite{balntas2017hpatches} for the homography estimation. HPatches contains 116 planar scenes with viewpoint or illumination changes, which is widely used to evaluate the low-level matching performance. Following~\cite{sarlin2020superglue,sun2021loftr}, we report the AUC of corner error up to thresholds 3, 5, and 10 pixels in Tab.~\ref{tab:hpatches}. RANSAC is adopted to get the robust homography matrix. To ensure fairness, we resize the short side of each image to 480 as LoFTR.
From Tab.~\ref{tab:hpatches}, CasMTRs trained on MegaDepth outperform other methods with denser matching results. But we should clarify that dense matching is not the key factor to improve the homography estimation. After the NMS detection, results from CasMTR are further improved with even fewer matches than LoFTR or QuadTree. Therefore, experiments on HPatches sufficiently show the effectiveness of the proposed method.
More details are discussed in the supplementary.

\begin{table}
\small 
\caption{Homography estimation on HPatches~\cite{balntas2017hpatches} with AUC of different corner errors (\%).\label{tab:hpatches}}
\vspace{-0.1in}
\centering
\begin{tabular}{c|ccc|c}
\hline
\multirow{2}{*}{Methods} & \multicolumn{3}{c|}{Pose Estimation AUC $\uparrow$} & \multirow{2}{*}{matches}\tabularnewline
\cline{2-4} \cline{3-4} \cline{4-4} 
 & @3px & @5px & @10px & \tabularnewline
\hline
DISK~\cite{tyszkiewicz2020disk}+NN & 52.3 & 64.9 & 78.9 & 1.1k\tabularnewline
SP~\cite{detone2018superpoint}+SuperGlue~\cite{sarlin2020superglue} & 53.9 & 68.3 & 81.7 & 0.6k\tabularnewline
LoFTR~\cite{sun2021loftr} & 64.6 & 74.8 & 84.2 & 2.6k\tabularnewline
QuadTree~\cite{tang2022quadtree} & 66.3 & 76.2 & 84.9 & 2.7k\tabularnewline
\hline
CasMTR-4c & 67.5 & 77.1 & 86.3 & 11.4k\tabularnewline
CasMTR-2c & 69.6 & 78.9 & 87.1 & 44.7k\tabularnewline
\hline
CasMTR-4c (NMS=5) & 69.7 & 78.8 & 87.0 & 0.4k\tabularnewline
CasMTR-2c (NMS=9) & \textbf{71.4} & \textbf{80.2} & \textbf{87.9} & 0.5k\tabularnewline
\hline
\end{tabular}
\vspace{-0.15in}
\end{table}

\subsection{Visual Localization}
We also evaluate CasMTR on the InLoc~\cite{taira2018inloc} and Aachen Day-Night v1.1~\cite{zhang2021reference} benchmarks of visual localization to further validate the robustness of our model. 
Following the pipeline of HLoc~\cite{sarlin2019coarse}, we replace the matching stage with compared methods for getting matching pairs between query and database images.
Since no official codes are provided from~\cite{sun2021loftr}, we re-implement the visual localization and report results in Tab.~\ref{tab:InLoc} and Tab.~\ref{tab:aachen}. Our baseline is based on QuadTree with Twins backbone. Considering high-resolution inputs and large-scale images, we use the MegaDepth pre-trained CasMTR-4c enhanced with NMS kernel size 5 to evaluate both benchmarks. As shown in Tab.~\ref{tab:InLoc} and Tab.~\ref{tab:aachen}, CasMTR outperforms other competitors. 

\begin{table}
\small 
\caption{Visual localization on InLoc~\cite{taira2018inloc}.  * means our implementation of LoFTR; note that our re-implementations are better on DUC1 and worse on DUC2 compared with~\cite{sun2021loftr}.
\label{tab:InLoc}}
\vspace{-0.15in}
\centering
\setlength{\tabcolsep}{0.6mm}{
\begin{tabular}{c|c|c}
\hline 
\multirow{2}{*}{Methods} & DUC1 & DUC2\tabularnewline
\cline{2-3} \cline{3-3} 
 & \multicolumn{2}{c}{(0.25m,2$^{\circ}$)/(0.5m,5$^{\circ}$)/(1m,10$^{\circ}$)}\tabularnewline
\hline 
HLoc~\cite{sarlin2019coarse}+LoFTR~\cite{sun2021loftr}{*} & 49.5/73.7/82.8 & \textbf{51.9}/69.5/80.9\tabularnewline
HLoc~\cite{sarlin2019coarse}+Baseline & 47.5/71.7/83.8 & 48.1/\textbf{70.2}/79.4\tabularnewline
HLoc~\cite{sarlin2019coarse}+CasMTR & \textbf{53.5}/\textbf{76.8}/\textbf{85.4} & \textbf{51.9}/\textbf{70.2}/\textbf{83.2}\tabularnewline
\hline 
\end{tabular}}
\vspace{-0.1in}
\end{table}

\begin{table}
\small 
\caption{Visual localization on Aachen Day-Night~\cite{zhang2021reference}.
\label{tab:aachen}}
\vspace{-0.15in}
\centering
\setlength{\tabcolsep}{0.7mm}{
\begin{tabular}{c|c|c}
\hline 
\multirow{2}{*}{Methods} & Day & Night\tabularnewline
\cline{2-3} \cline{3-3} 
 & \multicolumn{2}{c}{(0.25m,2$^{\circ}$) / (0.5m,5$^{\circ}$) / (1m,10$^{\circ}$)}\tabularnewline
\hline 
HLoc~\cite{sarlin2019coarse}+LoFTR~\cite{sun2021loftr} & 88.7/95.6/99.0 & \textbf{78.5}/90.6/99.0\tabularnewline
HLoc~\cite{sarlin2019coarse}+Aspanformer~\cite{chen2022aspanformer} & 89.4/95.6/99.0 & 77.5/\textbf{91.6}/\textbf{99.5}\tabularnewline
HLoc~\cite{sarlin2019coarse}+CasMTR & \textbf{90.4}/\textbf{96.2}/\textbf{99.3} & \textbf{78.5}/\textbf{91.6}/\textbf{99.5}\tabularnewline
\hline 
\end{tabular}}
\vspace{-0.1in}
\end{table}

\subsection{Ablations}
\label{sec:abaltions}

\begin{table}
\small 
\caption{Comparison between our CasMTR and the trivial refinement extension of baseline (Baseline-Tri) 
on MegaDepth. \label{tab:trivial_solution}}
\vspace{-0.15in}
\centering
\begin{tabular}{c|ccc}
\hline 
\multirow{2}{*}{Methods} & \multicolumn{3}{c}{Pose Estimation AUC $\uparrow$}\tabularnewline
\cline{2-4} \cline{3-4} \cline{4-4} 
 & @5$^{\circ}$ & @10$^{\circ}$ & @20$^{\circ}$\tabularnewline
\hline 
Baseline & 55.77 & 72.01 & 83.64\tabularnewline
\hline 
Baseline-Tri & 47.09 & 64.76 & 78.13\tabularnewline
Baseline-Tri (NMS=5) & 51.19 & 67.62 & 80.00\tabularnewline
Ours-4c (NMS=5) & 57.99 & 72.42 & 84.58\tabularnewline
Ours-2c (NMS=5) & \textbf{59.08} & \textbf{74.33} & \textbf{84.80}\tabularnewline
\hline 
\end{tabular}
\vspace{-0.1in}
\end{table}

\begin{figure*}
\begin{centering}
\includegraphics[width=0.92\linewidth]{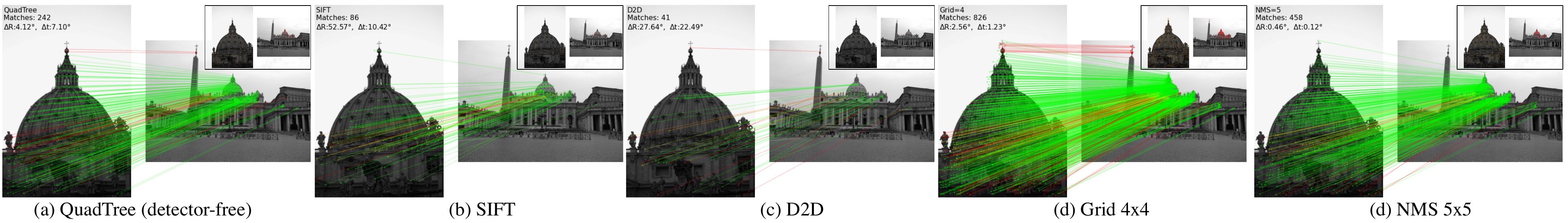} 
\par\end{centering}
\vspace{-0.1in}
\caption{Qualitative comparisons among different detection methods. Detected points are shown in the right-up corner..\label{fig:detection}}
\vspace{-0.15in}
\end{figure*}

\noindent\textbf{Cascade Matching vs Dense Refinement.}
As mentioned in Sec.~\ref{sec:refinement}, the straightforward way to achieve dense matching with detector-free methods~\cite{sun2021loftr,tang2022quadtree} is to make all patch-wise features in the refinement module produce matching results as much as possible. Theoretically, a such trivial extension can get as many matching pairs as CasMTR-2c. But as verified in Tab.~\ref{tab:trivial_solution}, such trivial extension (Baseline-Tri) fails to get good results. Because receptive fields of the patch-wise refinement are limited, while the self-attention only considers features in the same patch. Moreover, the cross-attention is also corrupted by non-overlapping cross windows. Although the NMS can improve the Baseline-Tri a little, it still has a large gap compared with CasMTR.



\begin{table}
\small 
\caption{Ablation studies about various detection methods based on CasMTR-4c trained on MegaDepth. Thr.$>$0.5 means that increasing the confidence threshold from 0.2 to 0.5. Trainable* and $\dag$ indicate that finetuning the baseline with extra trainable detectors. * only optimizes detected points while $\dag$ learns detected points with higher weights (3 times). Grid $4\times4$ means that selecting top-1 from non-overlapping $4\times4$ confidence windows, while NMS is maxpooled on $5\times5$ overlapping ones. \label{tab:ablation_detection}}
\vspace{-0.1in}
\centering
\begin{tabular}{c|ccc}
\hline 
\multirow{2}{*}{Detector} & \multicolumn{3}{c}{Pose Estimation AUC $\uparrow$}\tabularnewline
\cline{2-4} \cline{3-4} \cline{4-4} 
 & @5$^{\circ}$ & @10$^{\circ}$ & @20$^{\circ}$\tabularnewline
\hline 
-- & 56.34 & 72.11 & 83.55\tabularnewline
\hline 
Thr.$ >0.5$ & 53.81 & 70.46 & 82.54\tabularnewline
Trainable{*} & 56.06 & 72.05 & 83.26\tabularnewline
Trainable$\dag$ & 57.21 & 73.05 & 84.03\tabularnewline
SIFT & 51.84 & 68.78 & 81.39\tabularnewline
D2D & 53.92 & 70.50 & 82.58\tabularnewline
Grid 4$\times$4 & 56.54 & 72.42 & 83.98\tabularnewline
NMS 5$\times$5 & \textbf{57.99} & \textbf{73.36} & \textbf{84.58}\tabularnewline
\hline 
\end{tabular}
\vspace{-0.15in}
\end{table}

\noindent\textbf{Detection Methods.} 
We evaluate different detection strategies in Tab.~\ref{tab:ablation_detection}, while qualitative comparisons are shown in Fig.~\ref{fig:detection}.
Simply increasing the threshold is not effective for the pose estimation without considering the local relation. For trainable versions, we finetune CasMTR with another trainable detector jointly through the straight-through estimation~\cite{bengio2013estimating} with grid size $4\times4$ and tested with NMS kernel 5 as~\cite{tyszkiewicz2020disk}. But the trainable detector even reduces the performance. We think that such a jointly detect-and-describe pipeline is unsuitable for supervised image matching learning. Because the detector devotes to searching keypoints which are \textit{easy} to be matched, while the descriptor becomes lazy to ignore hard matching cases. Thus, we train another version by simply increasing the weights of detected points instead. This slightly improves the performance, but still has a gap from NMS. Note that NMS is much more efficient because it is training-free.
Moreover, both traditional SIFT~\cite{lowe2004distinctive} and feature-based D2D~\cite{tian2020d2d} post-processing fail to improve the matching performance. So these detectors are incompatible with the transformer-based matching, \textit{e.g.}, they cannot ensure that the detected keypoints enjoy confident model probabilities.
From Tab.~\ref{tab:ablation_detection}, the overlapping maxpool filtering (NMS) outperforms the non-overlapping one (Grid). 

\begin{table}
\small 
\caption{Ablation studies about NMS kernel size in post-processing (Post.) on MegaDepth.
\label{tab:ablation_nms}}
\vspace{-0.1in}
\centering
\begin{tabular}{cc|c|ccc}
\hline 
\multicolumn{2}{c|}{Cascade} & \multirow{2}{*}{Post.} & \multicolumn{3}{c}{Pose Estimation AUC $\uparrow$}\tabularnewline
\cline{1-2} \cline{2-2} \cline{4-6} \cline{5-6} \cline{6-6} 
4c & 2c &  & @5$^{\circ}$ & @10$^{\circ}$ & @20$^{\circ}$\tabularnewline
\hline 
 &  & -- & 55.77 & 72.01 & 83.64\tabularnewline
$\checkmark$ &  & -- & 56.34 & 72.11 & 83.55\tabularnewline
$\checkmark$ & $\checkmark$ & -- & 56.90 & 72.94 & 84.24\tabularnewline
\hline 
 &  & NMS 3$\times$3 & 56.23 & 72.17 & 83.37\tabularnewline
 &  & NMS 5$\times$5 & 55.41 & 71.10 & 82.67\tabularnewline
 &  & NMS 7$\times$7 & 55.75 & 71.09 & 82.26\tabularnewline
\hline 
$\checkmark$ &  & NMS 3$\times$3 & 56.95 & 73.02 & 84.36\tabularnewline
$\checkmark$ &  & NMS 5$\times$5 & 57.99 & 73.36 & 84.58\tabularnewline
$\checkmark$ &  & NMS 7$\times$7 & 56.99 & 72.78 & 84.11\tabularnewline
\hline 
$\checkmark$ & $\checkmark$ & NMS 3$\times$3 & 57.56 & 73.40 & 84.60\tabularnewline
$\checkmark$ & $\checkmark$ & NMS 5$\times$5 & \textbf{59.08} & \textbf{74.33} & \textbf{84.80}\tabularnewline
$\checkmark$ & $\checkmark$ & NMS 7$\times$7 & 57.64 & 73.44 & 84.38\tabularnewline
\hline 
\end{tabular}
\vspace{-0.15in}
\end{table}

\noindent\textbf{Kernel Size of NMS.}
We evaluate the NMS detection with different kernel sizes in Tab.~\ref{tab:ablation_nms} on MegaDepth 1152$\times$1152 image pairs. NMS can not improve the pose estimation of the baseline method without cascade dense matching. Because the semi-dense solution does not contain sufficient matching pairs to be detected as local keypoints. Besides, both CasMTR-4c and CasMTR-2c can achieve the best results with NMS kernel size 5. More experiments in HPatches (Tab.~\ref{tab:hpatches}) and visual localization (Tab.~\ref{tab:InLoc}, Tab.~\ref{tab:aachen}) denote that our NMS detection can work robustly with CasMTR.

\section{Conclusion}

We rethink the transformer-based image matching pipeline and find that locating spatially informative source points is critical. So we propose a transformer-based cascade matching model called CasMTR, which can produce denser matches compared with previous transformer-based methods through its coarse-to-fine cascade modules. Benefiting from the thorough investigation of efficient attention, CasMTR enjoys a good balance between performance and efficiency. Further, CasMTR can be finetuned based on off-the-shelf matching models through the PMT. 
The newly repurposed NMS further detects more precise matching pairs in informative keypoints, improving the pose estimation. 
CasMTR enjoys state-of-the-art results in relative pose estimation, homography estimation, and visual localization.

\noindent\textbf{Acknowledge.}
\small
Dr. Fu is also with Shanghai Key Lab of Intelligent Information Processing, Fudan University, and Fudan ISTBI—ZJNU Algorithm Centre for Brain-inspired Intelligence, Zhejiang Normal University, Jinhua, China.

{\small
\bibliographystyle{ieee_fullname}
\bibliography{egbib}

\begin{thebibliography}{10}\itemsep=-1pt

\bibitem{balntas2017hpatches}
Vassileios Balntas, Karel Lenc, Andrea Vedaldi, and Krystian Mikolajczyk.
\newblock Hpatches: A benchmark and evaluation of handcrafted and learned local
  descriptors.
\newblock In {\em Proceedings of the IEEE conference on computer vision and
  pattern recognition}, pages 5173--5182, 2017.

\bibitem{bengio2013estimating}
Yoshua Bengio, Nicholas L{\'e}onard, and Aaron Courville.
\newblock Estimating or propagating gradients through stochastic neurons for
  conditional computation.
\newblock {\em arXiv preprint arXiv:1308.3432}, 2013.

\bibitem{cao2022mvsformer}
Chenjie Cao, Xinlin Ren, and Yanwei Fu.
\newblock Mvsformer: Learning robust image representations via transformers and
  temperature-based depth for multi-view stereo.
\newblock {\em arXiv preprint arXiv:2208.02541}, 2022.

\bibitem{chen2022aspanformer}
Hongkai Chen, Zixin Luo, Lei Zhou, Yurun Tian, Mingmin Zhen, Tian Fang, David
  Mckinnon, Yanghai Tsin, and Long Quan.
\newblock Aspanformer: Detector-free image matching with adaptive span
  transformer.
\newblock {\em arXiv preprint arXiv:2208.14201}, 2022.

\bibitem{choy2016universal}
Christopher~B Choy, JunYoung Gwak, Silvio Savarese, and Manmohan Chandraker.
\newblock Universal correspondence network.
\newblock {\em Advances in neural information processing systems}, 29, 2016.

\bibitem{chu2021twins}
Xiangxiang Chu, Zhi Tian, Yuqing Wang, Bo Zhang, Haibing Ren, Xiaolin Wei,
  Huaxia Xia, and Chunhua Shen.
\newblock Twins: Revisiting the design of spatial attention in vision
  transformers.
\newblock {\em Advances in Neural Information Processing Systems},
  34:9355--9366, 2021.

\bibitem{dai2017scannet}
Angela Dai, Angel~X Chang, Manolis Savva, Maciej Halber, Thomas Funkhouser, and
  Matthias Nie{\ss}ner.
\newblock Scannet: Richly-annotated 3d reconstructions of indoor scenes.
\newblock In {\em Proceedings of the IEEE conference on computer vision and
  pattern recognition}, pages 5828--5839, 2017.

\bibitem{detone2017toward}
Daniel DeTone, Tomasz Malisiewicz, and Andrew Rabinovich.
\newblock Toward geometric deep slam.
\newblock {\em arXiv preprint arXiv:1707.07410}, 2017.

\bibitem{detone2018superpoint}
Daniel DeTone, Tomasz Malisiewicz, and Andrew Rabinovich.
\newblock Superpoint: Self-supervised interest point detection and description.
\newblock In {\em Proceedings of the IEEE conference on computer vision and
  pattern recognition workshops}, pages 224--236, 2018.

\bibitem{dong2022incremental}
Qiaole Dong, Chenjie Cao, and Yanwei Fu.
\newblock Incremental transformer structure enhanced image inpainting with
  masking positional encoding.
\newblock In {\em Proceedings of the IEEE/CVF Conference on Computer Vision and
  Pattern Recognition}, pages 11358--11368, 2022.

\bibitem{dusmanu2019d2}
Mihai Dusmanu, Ignacio Rocco, Tomas Pajdla, Marc Pollefeys, Josef Sivic,
  Akihiko Torii, and Torsten Sattler.
\newblock D2-net: A trainable cnn for joint description and detection of local
  features.
\newblock In {\em Proceedings of the ieee/cvf conference on computer vision and
  pattern recognition}, pages 8092--8101, 2019.

\bibitem{edstedt2022deep}
Johan Edstedt, M{\aa}rten Wadenb{\"a}ck, and Michael Felsberg.
\newblock Deep kernelized dense geometric matching.
\newblock {\em arXiv preprint arXiv:2202.00667}, 2022.

\bibitem{gu2020cascade}
Xiaodong Gu, Zhiwen Fan, Siyu Zhu, Zuozhuo Dai, Feitong Tan, and Ping Tan.
\newblock Cascade cost volume for high-resolution multi-view stereo and stereo
  matching.
\newblock In {\em Proceedings of the IEEE/CVF Conference on Computer Vision and
  Pattern Recognition}, pages 2495--2504, 2020.

\bibitem{guo2022visual}
Meng-Hao Guo, Cheng-Ze Lu, Zheng-Ning Liu, Ming-Ming Cheng, and Shi-Min Hu.
\newblock Visual attention network.
\newblock {\em arXiv preprint arXiv:2202.09741}, 2022.

\bibitem{han2015matchnet}
Xufeng Han, Thomas Leung, Yangqing Jia, Rahul Sukthankar, and Alexander~C Berg.
\newblock Matchnet: Unifying feature and metric learning for patch-based
  matching.
\newblock In {\em Proceedings of the IEEE conference on computer vision and
  pattern recognition}, pages 3279--3286, 2015.

\bibitem{houlsby2019parameter}
Neil Houlsby, Andrei Giurgiu, Stanislaw Jastrzebski, Bruna Morrone, Quentin
  De~Laroussilhe, Andrea Gesmundo, Mona Attariyan, and Sylvain Gelly.
\newblock Parameter-efficient transfer learning for nlp.
\newblock In {\em International Conference on Machine Learning}, pages
  2790--2799. PMLR, 2019.

\bibitem{huang2022flowformer}
Zhaoyang Huang, Xiaoyu Shi, Chao Zhang, Qiang Wang, Ka~Chun Cheung, Hongwei
  Qin, Jifeng Dai, and Hongsheng Li.
\newblock Flowformer: A transformer architecture for optical flow.
\newblock {\em arXiv preprint arXiv:2203.16194}, 2022.

\bibitem{jiang2021cotr}
Wei Jiang, Eduard Trulls, Jan Hosang, Andrea Tagliasacchi, and Kwang~Moo Yi.
\newblock Cotr: Correspondence transformer for matching across images.
\newblock In {\em Proceedings of the IEEE/CVF International Conference on
  Computer Vision}, pages 6207--6217, 2021.

\bibitem{katharopoulos2020transformers}
Angelos Katharopoulos, Apoorv Vyas, Nikolaos Pappas, and Fran{\c{c}}ois
  Fleuret.
\newblock Transformers are rnns: Fast autoregressive transformers with linear
  attention.
\newblock In {\em International Conference on Machine Learning}, pages
  5156--5165. PMLR, 2020.

\bibitem{li2020dual}
Xinghui Li, Kai Han, Shuda Li, and Victor Prisacariu.
\newblock Dual-resolution correspondence networks.
\newblock {\em Advances in Neural Information Processing Systems},
  33:17346--17357, 2020.

\bibitem{li2021prefix}
Xiang~Lisa Li and Percy Liang.
\newblock Prefix-tuning: Optimizing continuous prompts for generation.
\newblock {\em arXiv preprint arXiv:2101.00190}, 2021.

\bibitem{li2018megadepth}
Zhengqi Li and Noah Snavely.
\newblock Megadepth: Learning single-view depth prediction from internet
  photos.
\newblock In {\em Proceedings of the IEEE conference on computer vision and
  pattern recognition}, pages 2041--2050, 2018.

\bibitem{lin2017focal}
Tsung-Yi Lin, Priya Goyal, Ross Girshick, Kaiming He, and Piotr Doll{\'a}r.
\newblock Focal loss for dense object detection.
\newblock In {\em Proceedings of the IEEE international conference on computer
  vision}, pages 2980--2988, 2017.

\bibitem{liu2010sift}
Ce Liu, Jenny Yuen, and Antonio Torralba.
\newblock Sift flow: Dense correspondence across scenes and its applications.
\newblock {\em IEEE transactions on pattern analysis and machine intelligence},
  33(5):978--994, 2010.

\bibitem{liu2019gift}
Yuan Liu, Zehong Shen, Zhixuan Lin, Sida Peng, Hujun Bao, and Xiaowei Zhou.
\newblock Gift: Learning transformation-invariant dense visual descriptors via
  group cnns.
\newblock {\em Advances in Neural Information Processing Systems}, 32, 2019.

\bibitem{lowe2004distinctive}
David~G Lowe.
\newblock Distinctive image features from scale-invariant keypoints.
\newblock {\em International journal of computer vision}, 60(2):91--110, 2004.

\bibitem{luo2020aslfeat}
Zixin Luo, Lei Zhou, Xuyang Bai, Hongkai Chen, Jiahui Zhang, Yao Yao, Shiwei
  Li, Tian Fang, and Long Quan.
\newblock Aslfeat: Learning local features of accurate shape and localization.
\newblock In {\em Proceedings of the IEEE/CVF conference on computer vision and
  pattern recognition}, pages 6589--6598, 2020.

\bibitem{lynen2020large}
Simon Lynen, Bernhard Zeisl, Dror Aiger, Michael Bosse, Joel Hesch, Marc
  Pollefeys, Roland Siegwart, and Torsten Sattler.
\newblock Large-scale, real-time visual--inertial localization revisited.
\newblock {\em The International Journal of Robotics Research},
  39(9):1061--1084, 2020.

\bibitem{mi2022generalized}
Zhenxing Mi, Chang Di, and Dan Xu.
\newblock Generalized binary search network for highly-efficient multi-view
  stereo.
\newblock In {\em Proceedings of the IEEE/CVF Conference on Computer Vision and
  Pattern Recognition}, pages 12991--13000, 2022.

\bibitem{mishchuk2017working}
Anastasiia Mishchuk, Dmytro Mishkin, Filip Radenovic, and Jiri Matas.
\newblock Working hard to know your neighbor's margins: Local descriptor
  learning loss.
\newblock {\em Advances in neural information processing systems}, 30, 2017.

\bibitem{mur2015orb}
Raul Mur-Artal, Jose Maria~Martinez Montiel, and Juan~D Tardos.
\newblock Orb-slam: a versatile and accurate monocular slam system.
\newblock {\em IEEE transactions on robotics}, 31(5):1147--1163, 2015.

\bibitem{ranjan2017optical}
Anurag Ranjan and Michael~J Black.
\newblock Optical flow estimation using a spatial pyramid network.
\newblock In {\em Proceedings of the IEEE conference on computer vision and
  pattern recognition}, pages 4161--4170, 2017.

\bibitem{revaud2019r2d2}
Jerome Revaud, Philippe Weinzaepfel, C{\'e}sar De~Souza, Noe Pion, Gabriela
  Csurka, Yohann Cabon, and Martin Humenberger.
\newblock R2d2: repeatable and reliable detector and descriptor.
\newblock {\em arXiv preprint arXiv:1906.06195}, 2019.

\bibitem{rocco2020efficient}
Ignacio Rocco, Relja Arandjelovi{\'c}, and Josef Sivic.
\newblock Efficient neighbourhood consensus networks via submanifold sparse
  convolutions.
\newblock In {\em European conference on computer vision}, pages 605--621.
  Springer, 2020.

\bibitem{rocco2018neighbourhood}
Ignacio Rocco, Mircea Cimpoi, Relja Arandjelovi{\'c}, Akihiko Torii, Tomas
  Pajdla, and Josef Sivic.
\newblock Neighbourhood consensus networks.
\newblock {\em Advances in neural information processing systems}, 31, 2018.

\bibitem{rublee2011orb}
Ethan Rublee, Vincent Rabaud, Kurt Konolige, and Gary Bradski.
\newblock Orb: An efficient alternative to sift or surf.
\newblock In {\em 2011 International conference on computer vision}, pages
  2564--2571. Ieee, 2011.

\bibitem{sarlin2019coarse}
Paul-Edouard Sarlin, Cesar Cadena, Roland Siegwart, and Marcin Dymczyk.
\newblock From coarse to fine: Robust hierarchical localization at large scale.
\newblock In {\em Proceedings of the IEEE/CVF Conference on Computer Vision and
  Pattern Recognition}, pages 12716--12725, 2019.

\bibitem{sarlin2020superglue}
Paul-Edouard Sarlin, Daniel DeTone, Tomasz Malisiewicz, and Andrew Rabinovich.
\newblock Superglue: Learning feature matching with graph neural networks.
\newblock In {\em Proceedings of the IEEE/CVF conference on computer vision and
  pattern recognition}, pages 4938--4947, 2020.

\bibitem{schonberger2016structure}
Johannes~L Schonberger and Jan-Michael Frahm.
\newblock Structure-from-motion revisited.
\newblock In {\em Proceedings of the IEEE conference on computer vision and
  pattern recognition}, pages 4104--4113, 2016.

\bibitem{schonberger2016pixelwise}
Johannes~L Sch{\"o}nberger, Enliang Zheng, Jan-Michael Frahm, and Marc
  Pollefeys.
\newblock Pixelwise view selection for unstructured multi-view stereo.
\newblock In {\em European Conference on Computer Vision}, pages 501--518.
  Springer, 2016.

\bibitem{shi2022clustergnn}
Yan Shi, Jun-Xiong Cai, Yoli Shavit, Tai-Jiang Mu, Wensen Feng, and Kai Zhang.
\newblock Clustergnn: Cluster-based coarse-to-fine graph neural network for
  efficient feature matching.
\newblock In {\em Proceedings of the IEEE/CVF Conference on Computer Vision and
  Pattern Recognition}, pages 12517--12526, 2022.

\bibitem{sun2018pwc}
Deqing Sun, Xiaodong Yang, Ming-Yu Liu, and Jan Kautz.
\newblock Pwc-net: Cnns for optical flow using pyramid, warping, and cost
  volume.
\newblock In {\em Proceedings of the IEEE conference on computer vision and
  pattern recognition}, pages 8934--8943, 2018.

\bibitem{sun2021loftr}
Jiaming Sun, Zehong Shen, Yuang Wang, Hujun Bao, and Xiaowei Zhou.
\newblock Loftr: Detector-free local feature matching with transformers.
\newblock In {\em Proceedings of the IEEE/CVF conference on computer vision and
  pattern recognition}, pages 8922--8931, 2021.

\bibitem{sung2022lst}
Yi-Lin Sung, Jaemin Cho, and Mohit Bansal.
\newblock Lst: Ladder side-tuning for parameter and memory efficient transfer
  learning.
\newblock {\em arXiv preprint arXiv:2206.06522}, 2022.

\bibitem{taira2018inloc}
Hajime Taira, Masatoshi Okutomi, Torsten Sattler, Mircea Cimpoi, Marc
  Pollefeys, Josef Sivic, Tomas Pajdla, and Akihiko Torii.
\newblock Inloc: Indoor visual localization with dense matching and view
  synthesis.
\newblock In {\em Proceedings of the IEEE Conference on Computer Vision and
  Pattern Recognition}, pages 7199--7209, 2018.

\bibitem{tan2022eco}
Dongli Tan, Jiang-Jiang Liu, Xingyu Chen, Chao Chen, Ruixin Zhang, Yunhang
  Shen, Shouhong Ding, and Rongrong Ji.
\newblock Eco-tr: Efficient correspondences finding via coarse-to-fine
  refinement.
\newblock In {\em Computer Vision--ECCV 2022: 17th European Conference, Tel
  Aviv, Israel, October 23--27, 2022, Proceedings, Part X}, pages 317--334.
  Springer, 2022.

\bibitem{tang2022quadtree}
Shitao Tang, Jiahui Zhang, Siyu Zhu, and Ping Tan.
\newblock Quadtree attention for vision transformers.
\newblock {\em arXiv preprint arXiv:2201.02767}, 2022.

\bibitem{teed2020raft}
Zachary Teed and Jia Deng.
\newblock Raft: Recurrent all-pairs field transforms for optical flow.
\newblock In {\em European conference on computer vision}, pages 402--419.
  Springer, 2020.

\bibitem{tian2020d2d}
Yurun Tian, Vassileios Balntas, Tony Ng, Axel Barroso-Laguna, Yiannis Demiris,
  and Krystian Mikolajczyk.
\newblock D2d: Keypoint extraction with describe to detect approach.
\newblock In {\em Proceedings of the Asian Conference on Computer Vision},
  2020.

\bibitem{tian2019sosnet}
Yurun Tian, Xin Yu, Bin Fan, Fuchao Wu, Huub Heijnen, and Vassileios Balntas.
\newblock Sosnet: Second order similarity regularization for local descriptor
  learning.
\newblock In {\em Proceedings of the IEEE/CVF Conference on Computer Vision and
  Pattern Recognition}, pages 11016--11025, 2019.

\bibitem{tonioni2019real}
Alessio Tonioni, Fabio Tosi, Matteo Poggi, Stefano Mattoccia, and Luigi~Di
  Stefano.
\newblock Real-time self-adaptive deep stereo.
\newblock In {\em Proceedings of the IEEE/CVF Conference on Computer Vision and
  Pattern Recognition}, pages 195--204, 2019.

\bibitem{truong2020glu}
Prune Truong, Martin Danelljan, and Radu Timofte.
\newblock Glu-net: Global-local universal network for dense flow and
  correspondences.
\newblock In {\em Proceedings of the IEEE/CVF conference on computer vision and
  pattern recognition}, pages 6258--6268, 2020.

\bibitem{truong2021pdc}
Prune Truong, Martin Danelljan, Radu Timofte, and Luc Van~Gool.
\newblock Pdc-net+: Enhanced probabilistic dense correspondence network.
\newblock {\em arXiv preprint arXiv:2109.13912}, 2021.

\bibitem{truong2021learning}
Prune Truong, Martin Danelljan, Luc Van~Gool, and Radu Timofte.
\newblock Learning accurate dense correspondences and when to trust them.
\newblock In {\em Proceedings of the IEEE/CVF Conference on Computer Vision and
  Pattern Recognition}, pages 5714--5724, 2021.

\bibitem{tyszkiewicz2020disk}
Micha{\l} Tyszkiewicz, Pascal Fua, and Eduard Trulls.
\newblock Disk: Learning local features with policy gradient.
\newblock {\em Advances in Neural Information Processing Systems},
  33:14254--14265, 2020.

\bibitem{wang2021patchmatchnet}
Fangjinhua Wang, Silvano Galliani, Christoph Vogel, Pablo Speciale, and Marc
  Pollefeys.
\newblock Patchmatchnet: Learned multi-view patchmatch stereo.
\newblock In {\em Proceedings of the IEEE/CVF Conference on Computer Vision and
  Pattern Recognition}, pages 14194--14203, 2021.

\bibitem{wang2022matchformer}
Qing Wang, Jiaming Zhang, Kailun Yang, Kunyu Peng, and Rainer Stiefelhagen.
\newblock Matchformer: Interleaving attention in transformers for feature
  matching.
\newblock {\em arXiv preprint arXiv:2203.09645}, 2022.

\bibitem{wang2021pyramid}
Wenhai Wang, Enze Xie, Xiang Li, Deng-Ping Fan, Kaitao Song, Ding Liang, Tong
  Lu, Ping Luo, and Ling Shao.
\newblock Pyramid vision transformer: A versatile backbone for dense prediction
  without convolutions.
\newblock In {\em Proceedings of the IEEE/CVF International Conference on
  Computer Vision}, pages 568--578, 2021.

\bibitem{wang2019anytime}
Yan Wang, Zihang Lai, Gao Huang, Brian~H Wang, Laurens Van Der~Maaten, Mark
  Campbell, and Kilian~Q Weinberger.
\newblock Anytime stereo image depth estimation on mobile devices.
\newblock In {\em 2019 International Conference on Robotics and Automation
  (ICRA)}, pages 5893--5900. IEEE, 2019.

\bibitem{yang2019volumetric}
Gengshan Yang and Deva Ramanan.
\newblock Volumetric correspondence networks for optical flow.
\newblock {\em Advances in neural information processing systems}, 32, 2019.

\bibitem{yang2022non}
Jiayu Yang, Jose~M Alvarez, and Miaomiao Liu.
\newblock Non-parametric depth distribution modelling based depth inference for
  multi-view stereo.
\newblock In {\em Proceedings of the IEEE/CVF Conference on Computer Vision and
  Pattern Recognition}, pages 8626--8634, 2022.

\bibitem{yi2016lift}
Kwang~Moo Yi, Eduard Trulls, Vincent Lepetit, and Pascal Fua.
\newblock Lift: Learned invariant feature transform.
\newblock In {\em European conference on computer vision}, pages 467--483.
  Springer, 2016.

\bibitem{yin2019hierarchical}
Zhichao Yin, Trevor Darrell, and Fisher Yu.
\newblock Hierarchical discrete distribution decomposition for match density
  estimation.
\newblock In {\em Proceedings of the IEEE/CVF Conference on Computer Vision and
  Pattern Recognition}, pages 6044--6053, 2019.

\bibitem{zhang2020visibility}
Jingyang Zhang, Yao Yao, Shiwei Li, Zixin Luo, and Tian Fang.
\newblock Visibility-aware multi-view stereo network.
\newblock {\em arXiv preprint arXiv:2008.07928}, 2020.

\bibitem{zhang2021reference}
Zichao Zhang, Torsten Sattler, and Davide Scaramuzza.
\newblock Reference pose generation for long-term visual localization via
  learned features and view synthesis.
\newblock {\em International Journal of Computer Vision}, 129:821--844, 2021.

\bibitem{zhao2020maskflownet}
Shengyu Zhao, Yilun Sheng, Yue Dong, Eric~I Chang, Yan Xu, et~al.
\newblock Maskflownet: Asymmetric feature matching with learnable occlusion
  mask.
\newblock In {\em Proceedings of the IEEE/CVF Conference on Computer Vision and
  Pattern Recognition}, pages 6278--6287, 2020.

\bibitem{zhao2022global}
Shiyu Zhao, Long Zhao, Zhixing Zhang, Enyu Zhou, and Dimitris Metaxas.
\newblock Global matching with overlapping attention for optical flow
  estimation.
\newblock In {\em Proceedings of the IEEE/CVF Conference on Computer Vision and
  Pattern Recognition}, pages 17592--17601, 2022.

\bibitem{zhou2021patch2pix}
Qunjie Zhou, Torsten Sattler, and Laura Leal-Taixe.
\newblock Patch2pix: Epipolar-guided pixel-level correspondences.
\newblock In {\em Proceedings of the IEEE/CVF conference on computer vision and
  pattern recognition}, pages 4669--4678, 2021.

\end{thebibliography}
}

\appendix

\section{Implementation Details}

\noindent\textbf{Feature Encoder.} 
In this section, we provide more details about the implementation of CasMTR. 
As mentioned in the main paper, we replace partial layers of the feature encoder with pre-trained vision transformer layers from Twins-large~\cite{chu2021twins} for better generalization. Model channels are reduced to balance the computation. We list the detailed encoder architecture in Tab.~\ref{tab:feature_encoder}.
In particular, since the highest feature resolution of Twins is 1/4, we additionally train a CNN-based ResNet block for 1/2 features. Further, the 1/2 feature is combined to the encoder through the FPN.

\begin{table}[h]
\centering
\caption{Model details of feature encoders tackling features with different resolutions (Res.). `LSA+GSA' in our baseline indicate locally-grouped self-attention (LSA) and global sub-sampled attention (GSA) of Twins~\cite{chu2021twins}. Feature channels are listed in brackets.
\label{tab:feature_encoder}}
\small
\begin{tabular}{ccc}
\toprule
Feature Res. & Baseline & QuadTree\tabularnewline
\midrule
1/2 & ResBlock(64){*}2 & ResBlock(128){*}2\tabularnewline
1/4 & LSA+GSA(128){*}2 & ResBlock(196){*}2\tabularnewline
1/8 & LSA+GSA(256){*}2 & ResBlock(256){*}2\tabularnewline
\bottomrule
\end{tabular}
\end{table}

\noindent\textbf{Coarse and Cascade Matching Module.}
We almost follow QuadTree~\cite{tang2022quadtree} to design our coarse matching module. To relieve the computation from cascade modules, we reduce the attention block number in coarse stage from 8 to 6. As verified in our main paper, our CasMTR can still perform well with a slightly smaller coarse matching module.
For cascade modules, we use 4 and 3 attention blocks for 1/4 and 1/2 features respectively.
For 1/4 features, cascade modules are interlaced with `self-cross-self-cross' attention blocks, while cascade modules of 1/2 features are interlaced with `cross-self-cross' attention blocks. Since the self-attention is costly in high-resolution features, we tend to learn more cross-view information to make up the self one.

\noindent\textbf{Progressive Training on MegaDepth.}
CasMTR is trained progressively from scratch on MegaDepth~\cite{li2018megadepth}. 
To ensure that cascade modules can be optimized stably with enough valid matching pairs, \textit{i.e.}, one ground-truth match should appear in the local cascade searching space. We first train CasMTR with only $\frac{1}{8}$ resolution for 8 epochs, which can provide reliable initialization for the subsequent cascade learning. 
Based on the $\frac{1}{8}$ initialization, CasMTR-4c ($\frac{1}{4}$) is finetuned for 16 epochs while CasMTR-2c ($\frac{1}{4},\frac{1}{2}$) converges faster with 8 epochs. 
The 8-epoch training of $\frac{1}{8}$ CasMTR costs about 8 hours with batch size 16.
The 16-epoch training of CasMTR-4c costs about 1 day with batch size 8, while the 8-epoch training of CasMTR-2c costs about 2 days with batch size 4.
All training are based on 4 32GB V100 GPUs.

\noindent\textbf{Incremental Training on ScanNet.}
We adopt the PMT enhanced incremental tuning for CasMTR on ScanNet~\cite{dai2017scannet} based on the off-the-shelf QuadTree~\cite{tang2022quadtree} ScanNet model. The finetuning of PMT-CasMTR-4c is very efficient with only 2 epochs, which costs just 16 hours for batch size 32 with 4 48GB A6000 GPUs. Note that it would take about 5 days to re-train a competitive matching model on ScanNet in~\cite{chen2022aspanformer}. Moreover, our CasMTR-4c outperforms~\cite{chen2022aspanformer}.

\section{Self-Attention Modules}

\noindent\textbf{Linear Attention (Linear)~\cite{katharopoulos2020transformers}} works fast and efficiently for its quadratic \textit{channel} based complexity. It is also used as the standard attention in LoFTR~\cite{sun2021loftr}.

\noindent\textbf{Locally-grouped Self-Attention (LSA)~\cite{chu2021twins}} is simply learned within non-overlapping local patches while the window size is set in 7 as~\cite{chu2021twins}.

\noindent\textbf{Global Sub-sampled Attention (GSA)~\cite{wang2021pyramid}} downsamples keys and values to save the computation. LSA and GSA work complementarily in~\cite{chu2021twins} for both global and local feature learning. We set the downsample rate of GSA in 4.

\noindent\textbf{Simplified Top-k Attention (Top-k).} As the resolution raising, the QuadTree top-k attention~\cite{tang2022quadtree} becomes costly and infeasible. To overcome the heavy computation, here we adopt a simplified Top-k attention. To get the top-k keys and values for self-attention, we utilize the property of two-view matching. First, for each query, we match the top-1 patch from another image. Then, all matched top-1 patches search for top-k patches of original images through the coarse attention results, \textit{i.e.}, two-view cycle matching mentioned in the main paper. Therefore, all queries achieve their top-k keys and values without a global self-attention calculation. We set top-64 in this paper.

\noindent\textbf{Large Kernel Attention (LKA)~\cite{guo2022visual}} is built with pure convolutions, which comprises depthwise convolution (DW-Conv) and dilated DW-Conv. Compared with vanilla attention, LKA also enjoys large receptive fields, but it is less sensitive on the spatial scale. The effective LKA kernel size is 21.

\noindent\textbf{Patch-based OverLapping Attention (POLA)~\cite{zhao2022global}} can be seen as an extension of LSA. POLA magnifies receptive fields by employing overlapping windows with a larger window size for keys and values, while the query windows remain non-overlapping and have a small kernel size. Besides, POLA uses relative position encoding to further improve performance. Kernel sizes for query and key\&value are 7 and 21 in POLA respectively.

%

\section{Timing Analysis}
We have listed all time and memory costs in the main paper with different input scales. Here we further analyze each component's time cost of our CasMTR and QuadTree~\cite{tang2022quadtree} with an $832\times832$ image pair from MegaDepth in Tab.~\ref{tab:timing}. The testing is based on a 32GB V100 GPU.
From Tab.~\ref{tab:timing}, the transformer-based backbone in CasMTR just works slightly slower (8.75ms) than the CNN-based backbone in QuadTree. Since we reduce attention blocks in the coarse stage from 8 to 6, CasMTR costs less time for coarse attention. 
Moreover, our efficient cascade attention takes about 64.95ms and 146.26ms for 1/4 and 1/2 respectively. Compared to the 1/8 coarse attention of QuadTree, our cascade attention modules can tackle high-resolution features with 4 (1/4) and 16 (1/2) times sequence length. 
Furthermore, we should clarify that CasMTR-4c is good enough, which already achieves significant improvement on various downstream tasks with acceptable computation. 
Besides, NMS simplifies the final matching pairs; and vanilla attention works more efficiently than the linear one in the patch-wise refinement. Thus our refinement's time cost is still comparable to QuadTree's.

\noindent\textbf{Matching for Large Image Scales.}
CasMTR would not suffer from prohibitive running time even working for large-scale image matching as verified in Tab.~\ref{tab:timing2}. 
The testing is based on 1536$\times$1536 images on V100 GPU, while QuadTree can be seen as the baseline. 
As shown in the table, CasMTR is still comparable in high-resolution scenes.
Note that the proposed NMS filter could simplify matched pairs to less but more precise ones, which largely reduces the RANSAC time and further narrow the gap.
Besides, for extremely high-resolution images, CasMTR-2c is not necessary, while CasMTR-4c (matching in 1/4 resolution) is good enough, such as in-Loc results (most images $>$ 1300pix) in Tab.~\ref{tab:InLoc_supp}.

\begin{table}[h!]
\centering
\caption{Timing measurements of CasMTR and QuadTree with $832\times832$ image pairs. Res. means the resolution of feature maps in this module. 
\label{tab:timing}}
\small
\begin{tabular}{cccc}
\toprule
Process & Res. & QuadTree (ms) & CasMTR (ms)\tabularnewline
\midrule
Feature Extraction & -- & 44.37 & 53.12\tabularnewline
Coarse Attention & 1/8 & 125.62 & 111.38\tabularnewline
Coarse Matching & 1/8 & 23.73 & 29.05\tabularnewline
\midrule
Cascade Attention & 1/4 & -- & 64.95\tabularnewline
Cascade Matching & 1/4 & -- & 8.67\tabularnewline
\midrule
Cascade Attention & 1/2 & -- & 146.26\tabularnewline
Cascade Matching & 1/2 & -- & 18.91\tabularnewline
\midrule
Refinement & 1/2 & 9.14 & 10.80\tabularnewline
\midrule
Total & -- & 202.86 & 443.14\tabularnewline
\bottomrule
\end{tabular}
\end{table}

\begin{table}[h!]
\centering
\caption{Inference cost (sec/pair) on 1536$\times$1536 image pairs compared with QuadTree~\cite{tang2022quadtree} and CasMTR.
\label{tab:timing2}}
\vspace{-0.1in}
\small
\begin{tabular}{cccc}
\toprule
 & QuadTree & CasMTR-4c & CasMTR-2c \\
\midrule
Matching & 1.25 & 1.49 (19\%$\uparrow$) & 1.82 (46\%$\uparrow$) \\
+RANSAC & 1.37 & 1.54 (12\%$\uparrow$) & 1.87 (36\%$\uparrow$) \\
\bottomrule 
\end{tabular}
\end{table}

\section{Supplemental Experiments}

\subsection{Ablations about Cascade Scales}
We compare the CasMTR with different cascade scales on MegaDepth in Tab.~\ref{tab:ablation_cas}. The baseline is the first row with only a coarse stage in 1/8. Note that when we extend our model with a more coarse initialization (1/16), CasMTR cannot achieve proper matching results. Although starting from the 1/16 coarse matching is more efficient, matching in 1/16 features is too challenging and causes inevitable errors that corrupt subsequent cascade learning.

\begin{table}[h]
\centering
\caption{Ablation studies of the cascade scale without post-processing on MegaDepth. \label{tab:ablation_cas}}
\small 
\begin{tabular}{c|c|ccc}
\hline 
\multirow{2}{*}{Coarse} & \multirow{2}{*}{Cascade} & \multicolumn{3}{c}{Pose Estimation AUC $\uparrow$}\tabularnewline
\cline{3-5} \cline{4-5} \cline{5-5} 
 &  & @5$^{\circ}$ & @10$^{\circ}$ & @20$^{\circ}$\tabularnewline
\hline 
1/8 & -- & 55.77 & 72.01 & 83.64\tabularnewline
\hline 
1/8 & 1/4 & 56.34 & 72.11 & 83.55\tabularnewline
1/16 & 1/8, 1/4 & 49.26 & 66.27 & 79.40\tabularnewline
1/8 & 1/4, 1/2 & \textbf{56.90} & \textbf{72.94} & \textbf{84.24}\tabularnewline
\hline 
\end{tabular}
\end{table}

\subsection{Ablations about PMT}
We compare the CasMTR with and without the Parameter and Memory-efficient Tuning (PMT) on ScanNet in Tab.~\ref{tab:ablation_lst}. While without the PMT, CasMTR-4c directly utilizes frozen FPN features from the off-the-shelf QuadTree matching model~\cite{tang2022quadtree}. CasMTR based on PMT outperforms the one without it. PMT enjoys both parameter and memory efficiency with only 0.97M trainable parameters, while the whole FPN has 5.91M trainable ones. Note that we dose not try to finetune the QuadTree FPN, because it will disturb the coarse matching initialization; and training the whole QuadTree model with coarse attention modules is very costly and unnecessary for our incremental training.

\begin{table}[h]
\centering
\caption{Ablations of CasMTR-4c with/without PMT on ScanNet. \label{tab:ablation_lst}}
\small 
\begin{tabular}{c|ccc}
\hline 
\multirow{2}{*}{PMT} & \multicolumn{3}{c}{Pose Estimation AUC $\uparrow$}\tabularnewline
\cline{2-4} \cline{3-4} \cline{4-4} 
 & @5$^{\circ}$ & @10$^{\circ}$ & @20$^{\circ}$\tabularnewline
\hline 
$\checkmark$ & \textbf{27.1} & \textbf{47.0} & \textbf{64.4}\tabularnewline
\hline 
$\times$ & 26.2 & 46.1 & 63.5\tabularnewline
\hline 
\end{tabular}
\end{table}

\subsection{Ablations about NMS on HPatches}

We further compare CasMTR with different NMS kernel sizes on HPatches~\cite{balntas2017hpatches} for the homography estimation in Tab.~\ref{tab:nms_hpatches}.
First, both CasMTR-4c and CasMTR-2c without NMS outperform QuadTree. Especially, NMS kernels 5 and 9 perform best for CasMTR-4c and CasMTR-2c respectively. And our methods achieve impressive results with only 400 to 500 averaged matches, which are much fewer than QuadTree's.
We think that the keypoint location is important for homography estimation, while the proposed NMS detection can work properly for it. Note that the setting of NMS kernel 5 of the relative pose estimation is still competitive on HPatches, which shows a good generalization of NMS.

\begin{table}[h]
\centering
\caption{Ablation studies about NMS kernel size (k) on HPatches. \label{tab:nms_hpatches}}
\small 
\begin{tabular}{c|c|ccc|c}
\hline 
\multirow{2}{*}{Methods} & \multirow{2}{*}{NMS} & \multicolumn{3}{c|}{Pose Estimation AUC} & \multirow{2}{*}{matches}\tabularnewline
\cline{3-5} \cline{4-5} \cline{5-5} 
 &  & @3px & @5px & @10px & \tabularnewline
\hline 
QuadTree & -- & 66.37 & 76.23 & 84.97 & 2749\tabularnewline
\hline 
CasMTR-4c & -- & 67.50 & 77.10 & 86.25 & 11439\tabularnewline
CasMTR-4c & k=3 & 67.71 & 77.45 & 86.16 & 923\tabularnewline
CasMTR-4c & k=5 & \textbf{69.71} & \textbf{78.81} & 86.96 & 400\tabularnewline
CasMTR-4c & k=7 & 69.70 & 78.76 & \textbf{87.01} & 212\tabularnewline
\hline 
CasMTR-2c & -- & 69.06 & 78.47 & 86.75 & 44754\tabularnewline
CasMTR-2c & k=3 & 70.11 & 79.15 & 87.30 & 3520\tabularnewline
CasMTR-2c & k=5 & 70.35 & 79.60 & 87.59 & 1477\tabularnewline
CasMTR-2c & k=7 & 70.89 & 79.68 & 87.73 & 778\tabularnewline
CasMTR-2c & k=9 & \textbf{71.43} & \textbf{80.20} & \textbf{87.91} & 507\tabularnewline
CasMTR-2c & k=11 & 71.19 & 79.92 & 87.69 & 351\tabularnewline
\hline 
\end{tabular}
\end{table}

\subsection{Qualitative Ablations about NMS Kernel}

\begin{figure*}
\begin{centering}
\includegraphics[width=0.95\linewidth]{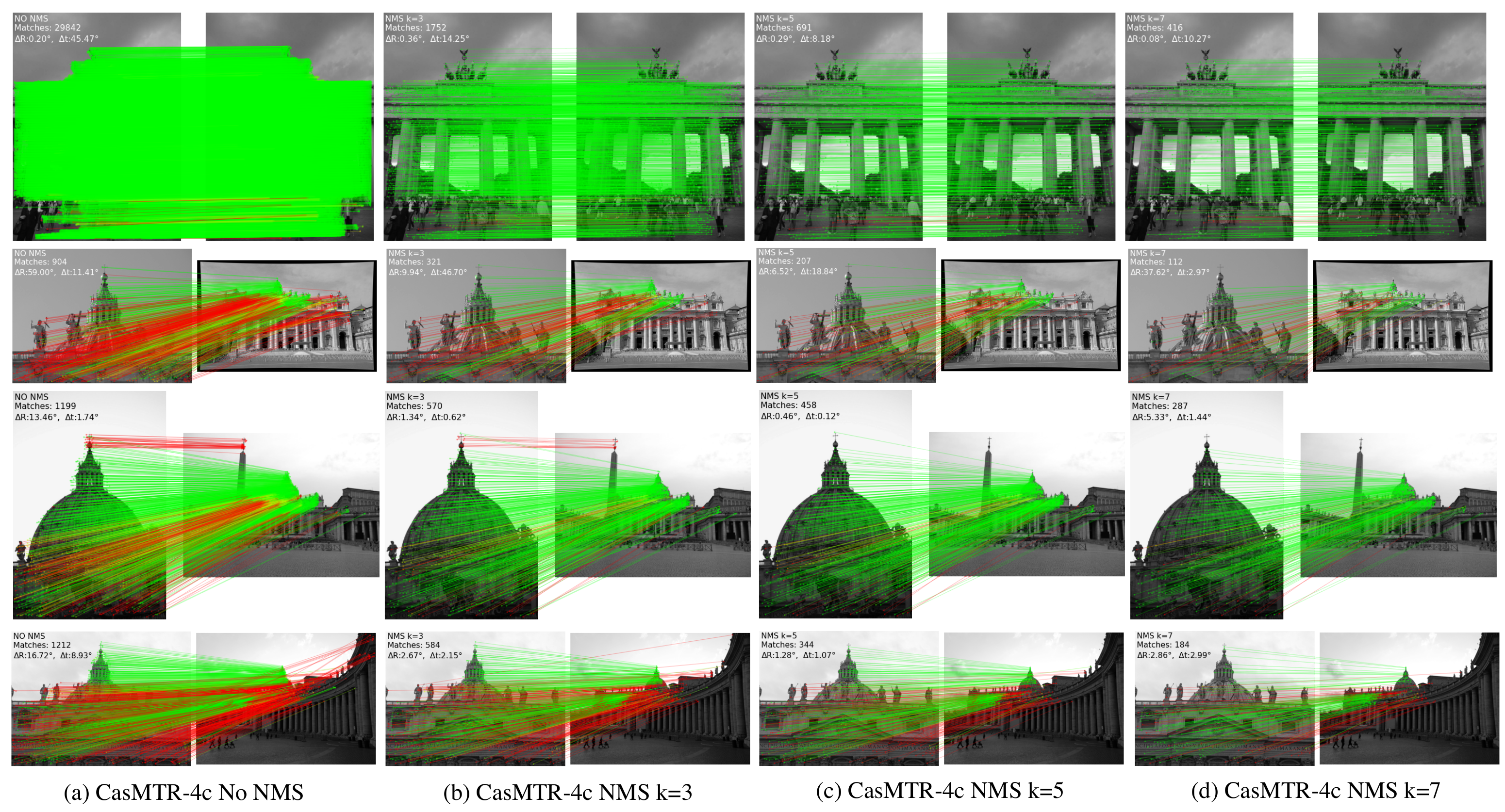} 
\par\end{centering}
\caption{Qualitative comparisons of CasMTR-4c among different NMS kernels on MegaDepth. Please zoom-in for details.\label{fig:nms_quali}}
\end{figure*}

We show some qualitative samples of CasMTR-4c with different NMS kernels in Fig.~\ref{fig:nms_quali}. We find that NMS detection is very effective for two types of image pairs. The first type is image pairs with large displacements (rows 2,3,4 of Fig.~\ref{fig:nms_quali}). These pairs are difficult for matching algorithms to achieve precise matches; and denser predictions usually cause more incorrect matches, which discourage the performance. So detecting local informative keypoints through the NMS is critical for improvement. On the other hand, image pairs with very limited displacements are also challenging (row 1 of Fig.~\ref{fig:nms_quali}), which may cause large translation errors. For these image pairs, NMS detection is also useful to achieve superior performance with more accurate keypoint matches.

\subsection{More Ablation Studies on InLoc}

We compare CasMTR-4c and CasMTR-2c with/without NMS (kernel size 5) on InLoc~\cite{taira2018inloc} in Tab.~\ref{tab:InLoc_supp}. And we did not further tune the kernel size of NMS on InLoc to ensure the fairness. CasMTR-2c achieves better results than CasMTR-4c on DUC1, but it performs worse on DUC2. Taking the trade-off of efficiency and performance into consideration, we adopt CasMTR-4c as our final solution in the main paper. 

\begin{table}
\small 
\caption{Visual localization results on InLoc~\cite{taira2018inloc}.  * means our implementation of LoFTR; note that results of our implementation are better on DUC1 and worse on DUC2 than those reported in~\cite{sun2021loftr}. 
\label{tab:InLoc_supp}}
\centering
\setlength{\tabcolsep}{0.7mm}{
\begin{tabular}{c|c|c}
\hline 
\multirow{2}{*}{Methods} & DUC1 & DUC2\tabularnewline
\cline{2-3} \cline{3-3} 
 & \multicolumn{2}{c}{(0.25m,2$^{\circ}$) / (0.5m,5$^{\circ}$) / (1m,10$^{\circ}$)}\tabularnewline
\hline 
HLoc~\cite{sarlin2019coarse}+LoFTR~\cite{sun2021loftr}{*} & 49.5/73.7/82.8 & \textbf{51.9}/69.5/80.9\tabularnewline
HLoc~\cite{sarlin2019coarse}+Baseline & 47.5/71.7/83.8 & 48.1/\textbf{70.2}/79.4\tabularnewline
HLoc~\cite{sarlin2019coarse}+CasMTR-4c & 51.0/75.3/84.3 & 51.1/69.5/\textbf{83.2}\tabularnewline
HLoc~\cite{sarlin2019coarse}+CasMTR-2c & 50.5/74.7/83.8 & 49.6/\textbf{70.2}/\textbf{83.2}\tabularnewline
HLoc~\cite{sarlin2019coarse}+CasMTR-4c-NMS5 & \textbf{53.5}/76.8/85.4 & \textbf{51.9}/\textbf{70.2}/\textbf{83.2}\tabularnewline
HLoc~\cite{sarlin2019coarse}+CasMTR-2c-NMS5 & 53.0/\textbf{77.8}/\textbf{86.4} & 49.6/\textbf{70.2}/82.4\tabularnewline
\hline 
\end{tabular}}
\end{table}

\subsection{Matching in Extremely Low Resolutions}
Learning the capability for extremely low-resolution matching is interesting because we could not always guarantee access to high-quality images. Results are shown in Tab.~\ref{tab:LR-exp}. We further compare results from SuperPoint~\cite{detone2018superpoint}+SuperGlue~\cite{sarlin2020superglue}.
The performances of both detector-based and detector-free methods are dramatically degraded in extremely low-resolution matching. However, our CasMTRs enjoy better robustness. As the resolution is reduced, the advantages of our algorithm become more apparent, especially for the CasMTR-2c. Note that for 256$\times$256, the matching is very challenging; and our method enjoys about 120\%, 82\%, and 50\% improvements on AUC5$^\circ$, AUC10$^\circ$, and AUC20$^\circ$ compared to QuadTree.

\begin{table*}[h!]
\small 
\caption{Matching for image pairs with extremely low resolutions.
\label{tab:LR-exp}}
\centering
\begin{tabular}{c|c|c|c|c|c|c|c|c|c}
\hline 
\multirow{2}{*}{Methods} & \multicolumn{3}{c|}{640$\times$640} & \multicolumn{3}{c|}{512$\times$512} & \multicolumn{3}{c}{256$\times$256}\tabularnewline
\cline{2-10} \cline{3-10} \cline{4-10} \cline{5-10} \cline{6-10} \cline{7-10} \cline{8-10} \cline{9-10} \cline{10-10} 
 & AUC5 & AUC10 & AUC20 & AUC5 & AUC10 & AUC20 & AUC5 & AUC10 & AUC20\tabularnewline
\hline 
\hline 
QuadTree & 49.86 & 66.85 & 79.43 & 44.06 & 61.35 & 75.15 & 10.42 & 22.04 & 38.22\tabularnewline
SuperGlue & 27.55 & 44.43 & 61.63 & 18.46 & 33.60 & 51.31 & 1.64 & 5.11 & 13.59\tabularnewline
\hline 
CasMTR-4c & 51.11 & 67.76 & 80.49 & 47.07 & 64.21 & 77.81 & 12.70 & 26.77 & 44.64\tabularnewline
CasMTR-2c & \textbf{54.98} & \textbf{71.48} & \textbf{83.11} & \textbf{51.36} & \textbf{68.08} & \textbf{80.78} & \textbf{23.38} & \textbf{40.24} & \textbf{57.11}\tabularnewline
\hline 
\end{tabular}
\end{table*}

\subsection{Insights about CasMTR}

\begin{figure*}
\begin{centering}
\includegraphics[width=0.95\linewidth]{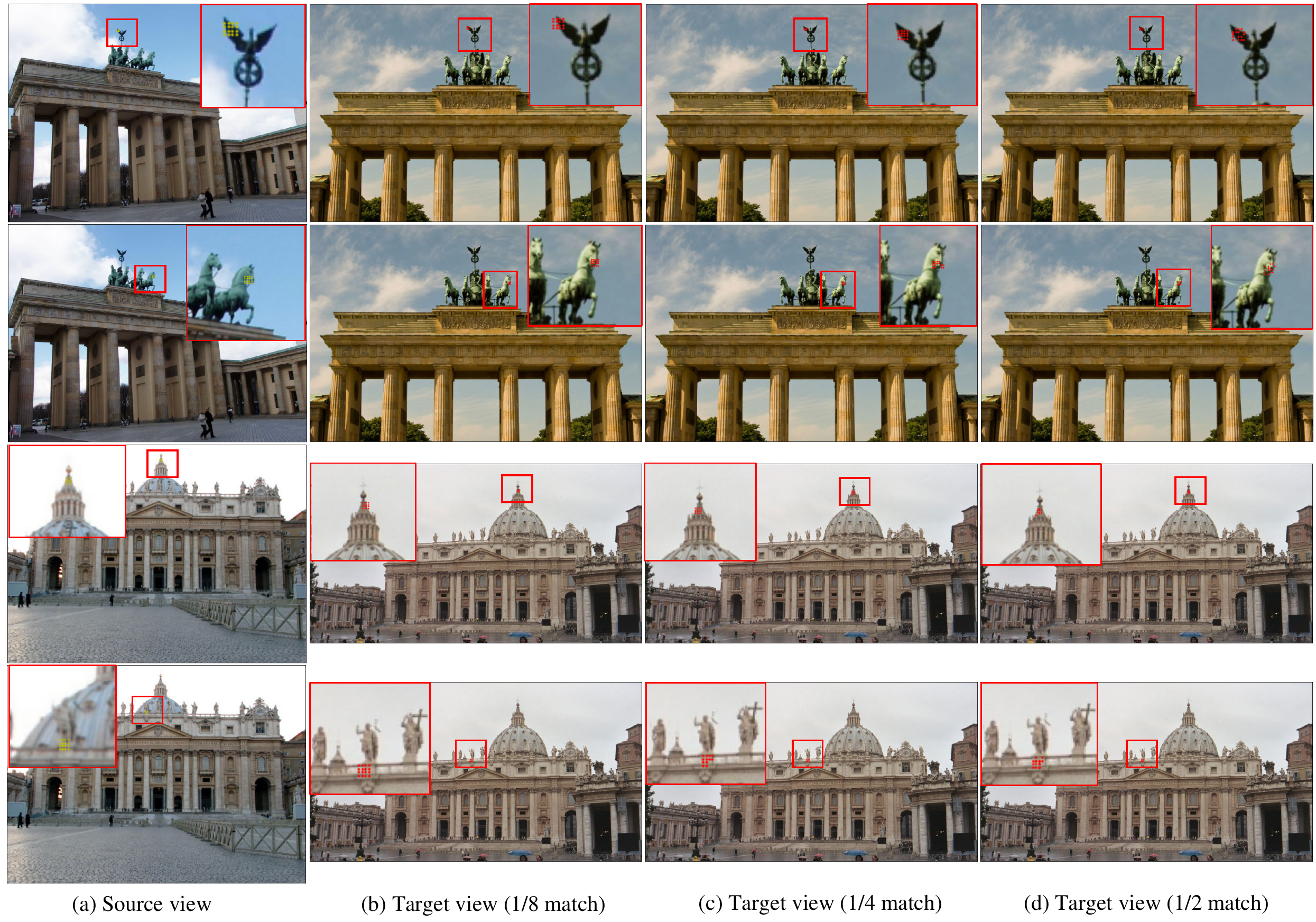} 
\par\end{centering}
\caption{The effectiveness of each cascade stage from CasMTR. Cascade modules can correctly refine the dense matching results rather than trivially searching for the nearest neighbor. Please zoom-in for details.\label{fig:insights}}
\end{figure*}

In this section we further discuss some additional insights about the proposed CasMTR. As shown in Fig.~\ref{fig:insights}, matching in the coarse stage (1/8) usually suffers from some inevitable deviations.
In particular, large displacements of viewpoint and occlusions break the rule that a local patch in source view (yellow points in Fig.~\ref{fig:insights}(a)) should be matched to a patch with the same size in target view (red points in Fig.~\ref{fig:insights}(b)).
Rather than trivially searching for the nearest neighbor, our CasMTR can gradually refine all matched points of the target view to more exact locations. Thus CasMTR further improves the pose estimation with lower detailed error.

\subsection{More Qualitative Comparisons}

\begin{figure*}[h!]
\begin{centering}
\includegraphics[width=0.9\linewidth]{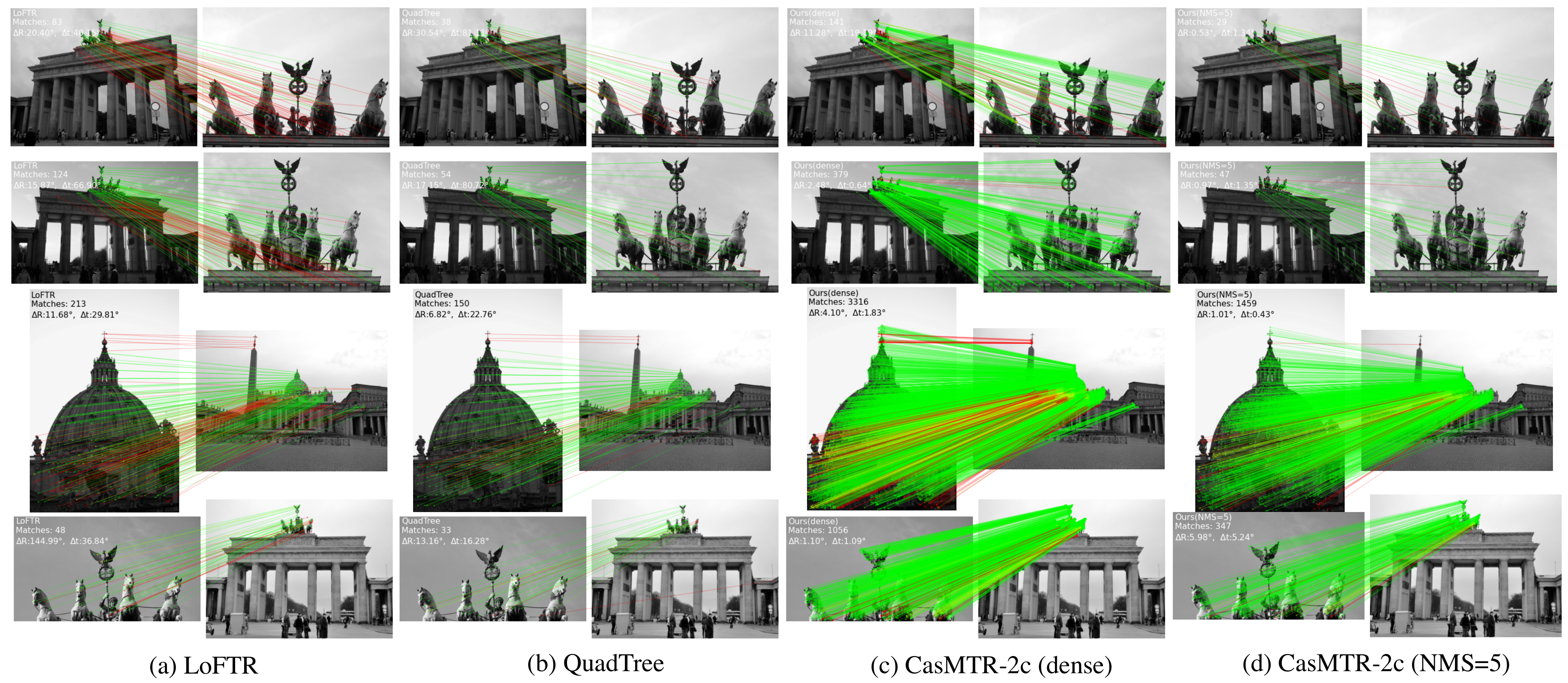} 
\par\end{centering}
\caption{Qualitative results compared on MegaDepth~\cite{li2018megadepth}. Please zoom-in for details.\label{fig:magadepth_supp}}
\end{figure*}

\begin{figure*}[h!]
\begin{centering}
\includegraphics[width=0.95\linewidth]{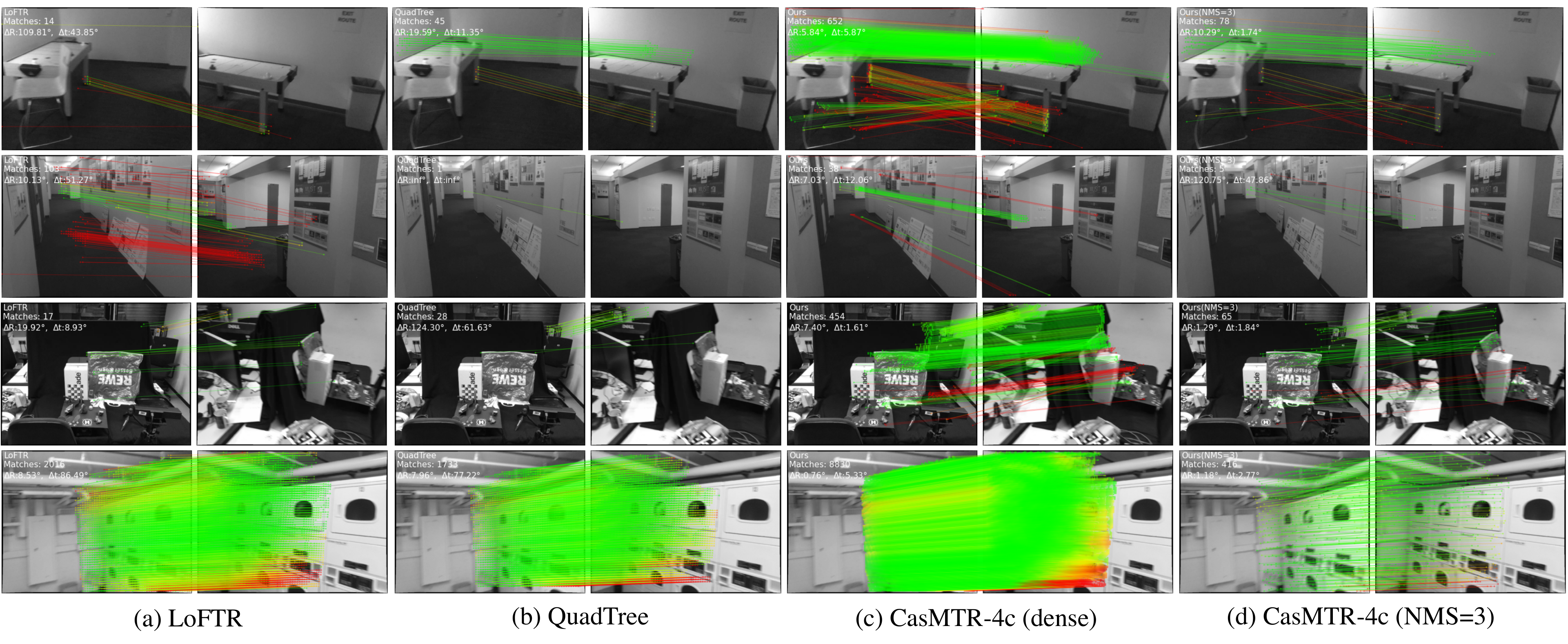} 
\par\end{centering}
\vspace{-0.15in}
\caption{Qualitative results compared on ScanNet~\cite{dai2017scannet}. Please zoom-in for details.\label{fig:scannet_supp}}
\vspace{-0.15in}
\end{figure*}

\begin{figure*}[h!]
\begin{centering}
\includegraphics[width=0.95\linewidth]{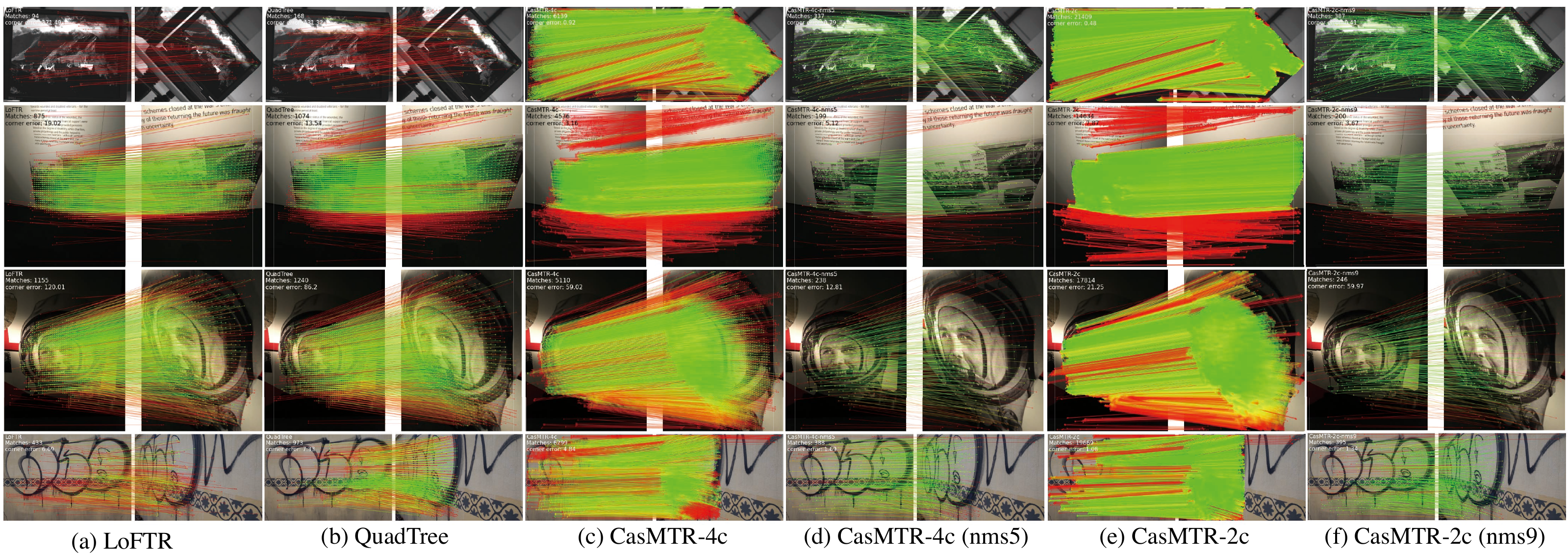} 
\par\end{centering}
\vspace{-0.15in}
\caption{Qualitative results compared on HPatches~\cite{balntas2017hpatches}. All image pairs are resized to meet that the short side is 480. We also show the corner error of each instance. The matching color threshold is 2-pixel. Please zoom-in for details.\label{fig:hpatches_supp}}
\vspace{-0.15in}
\end{figure*}

\begin{figure*}[h!]
\begin{centering}
\includegraphics[width=0.95\linewidth]{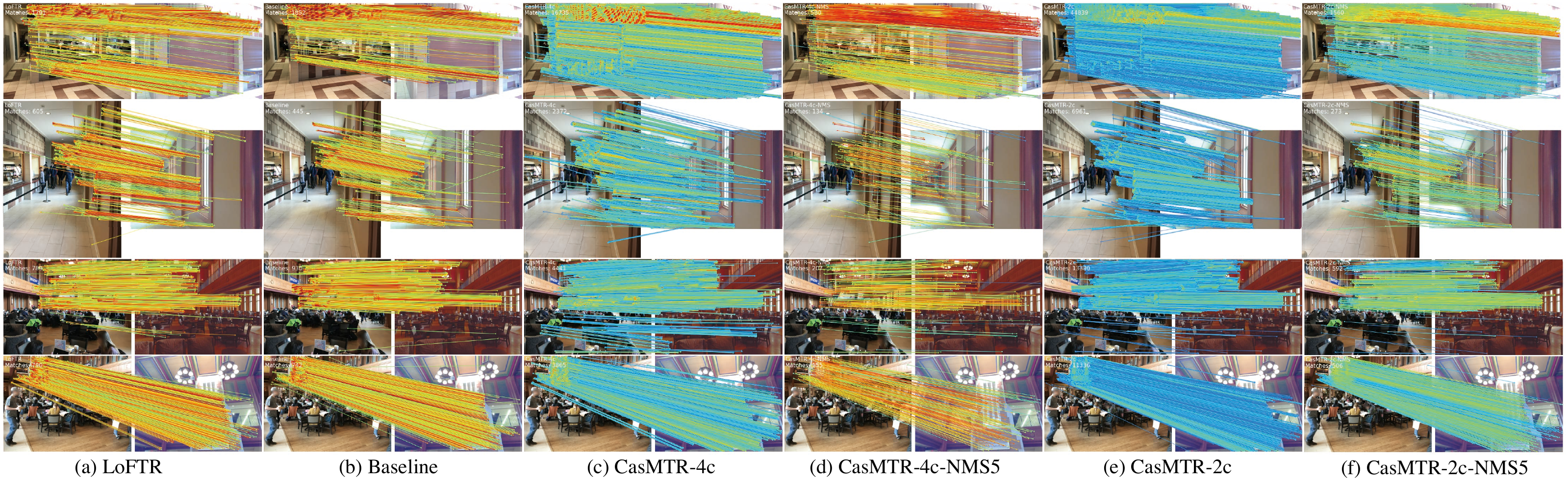} 
\par\end{centering}
\vspace{-0.15in}
\caption{Qualitative results compared on InLoc~\cite{taira2018inloc}. All image pairs are resized to meet that the short side is 1024. We colorize the matches with model confidence. Red means certain while blue means uncertain. Please zoom-in for details.\label{fig:inloc_supp}}
\vspace{-0.15in}
\end{figure*}

More qualitative comparisons for MegaDepth~\cite{li2018megadepth} and ScanNet~\cite{dai2017scannet} are shown in Fig.~\ref{fig:magadepth_supp} and Fig.~\ref{fig:scannet_supp} respectively.
From Fig.~\ref{fig:magadepth_supp}, CasMTR-2c outperforms LoFTR~\cite{sun2021loftr} and QuadTree~\cite{tang2022quadtree}, while NMS can further improve the results.
From Fig.~\ref{fig:scannet_supp}, we find that denser matches without NMS are more suitable for ScanNet images with textureless regions and limited resolutions. 
Besides, note that some sparse matches are filtered by the NMS (row2 of Fig.~\ref{fig:scannet_supp}). Because our NMS is only based on the densest confidence map (1/4 of ScanNet). Thus keypoints detected by the NMS may have low-confident scores in the frozen coarse stage, so these points would be eliminated by the confidence threshold (all stages' confidence thresholds in our work are fixed in 0.2). 
Moreover, qualitative results on HPatches~\cite{balntas2017hpatches} are shown in Fig.~\ref{fig:hpatches_supp}. And we also provided some results from InLoc~\cite{taira2018inloc} of the visual localization task in Fig.~\ref{fig:inloc_supp}. Note that the ground truth of InLoc is not provided. So we colorize the matches with model confidence. It seems that our results of Fig.~\ref{fig:inloc_supp}(c,e) have lower confidence. Because the results of CasMTR are much denser than LoFTR and baseline. Thus many low-confident matches are remained to cover the high-confident ones. Moreover, our NMS can successfully detect keypoints with locally high confidence, which improves the performance as in Tab.~\ref{tab:InLoc_supp}.

\section{Limitation}
The proposed CasMTR can achieve good performance in various matching downstream tasks. Although CasMTR-4c with 1/4 feature maps enjoys impressive enough results, CasMTR-2c with 1/2 high-resolution features can further improve the results in most situations. So we think that learning high-resolution attention modules is still necessary for image matching. Though we have made lots of efforts to improve the efficiency, learning in 1/2 features is still very challenging as shown in Tab.~\ref{tab:timing}. Therefore, improving the efficiency of the high-resolution feature correlation learning for attention modules should be an interesting future work.
On the other hand, NMS fails to be generalized well on texture-less indoor scenes with a frozen coarse stage. Therefore, we consider it as future work to integrate both trainable and un-trainable confidence for NMS detection in challenging scenes.

\end{document}